\documentclass[journal]{IEEEtran}
\usepackage{amssymb,multirow,afterpage}
\ifCLASSINFOpdf
  \usepackage[pdftex]{graphicx}
\else
  \usepackage[dvips]{graphicx}
\fi
\usepackage[cmex10]{amsmath}
\usepackage{algpseudocode}

\algrenewcommand\textproc{}

\ifCLASSOPTIONcompsoc
 \usepackage[caption=false,font=normalsize,labelfont=sf,textfont=sf]{subfig}
\else
 \usepackage[caption=false,font=footnotesize]{subfig}
\fi
\usepackage{fixltx2e}
\usepackage{url}

\begin{document}
%
\title{Image Segmentation Using Hierarchical Merge Tree}
%
%
%

\author{
Ting~Liu, Mojtaba~Seyedhosseini, and~Tolga~Tasdizen,~\IEEEmembership{Senior~Member,~IEEE}
\thanks{T. Liu, M. Seyedhosseini, and T. Tasdizen are with the Scientific Computing and Imaging Institute, University of Utah, Salt Lake City, UT 84112, USA.
E-mail: \{ting,mseyed,tolga\}@sci.utah.edu}
\thanks{T. Liu is also with the School of Computing, University of Utah.}
\thanks{M. Seyedhosseini and T. Tasdizen are also with the Department of Electrical and Computer Engineering, University of Utah.}
}

\maketitle

\begin{abstract}
This paper investigates one of the most fundamental computer vision problems: image segmentation. We propose a supervised hierarchical approach to object-independent image segmentation. Starting with over-segmenting superpixels, we use a tree structure to represent the hierarchy of region merging, by which we reduce the problem of segmenting image regions to finding a set of label assignment to tree nodes. We formulate the tree structure as a constrained conditional model to associate region merging with likelihoods predicted using an ensemble boundary classifier. Final segmentations can then be inferred by finding globally optimal solutions to the model efficiently. We also present an iterative training and testing algorithm that generates various tree structures and combines them to emphasize accurate boundaries by segmentation accumulation. Experiment results and comparisons with other recent methods on six public data sets demonstrate that our approach achieves state-of-the-art region accuracy and is competitive in image segmentation without semantic priors.
\end{abstract}


%
\IEEEpeerreviewmaketitle

\section{Introduction}
%
%
%
%
\IEEEPARstart{I}{mage} segmentation is an important mid-level computer vision problem that has been studied for a long time yet remains challenging. General image segmentation is used as a pre-processing step for solving high-level vision problems, such as object recognition and image classification. In many inter-disciplinary areas, e.g., biological and medical imaging, image segmentation also plays a significant role in helping scientists quantify and analyze image data. While a lot of research has been done to achieve high segmentation accuracy for specific types of images, the quality of image segmentation for general scenes is still less than satisfactory.

In this paper, we introduce a supervised learning based image segmentation framework, namely, the hierarchical merge tree model. Starting with over-segmenting superpixels, we propose to represent the region merging hierarchy with a tree-like constrained conditional model. An ensemble boundary classifier is trained to score each factor in the graphical model. A globally optimal label assignment to the model is computed by minimizing the total energy under the region consistency constraint, and a final segmentation is recovered from the labeling. We also propose an iterative approach that generates various region merging hierarchies and combines them to improve the overall performance via segmentation accumulation. We conduct extensive experiments for empirical validation. Comparisons with other very recent methods on six public data sets show that our proposed method produces state-of-the-art region segmentation results.

We begin with a review of previous related work on general image segmentation methods in Section~\ref{sec:previous_work}. In Section~\ref{sec:hierarchical_merge_tree_model}, we illustrate our hierarchical merge tree as a constrained conditional model and introduce the ensemble boundary classifier. In Section~\ref{sec:iterative_hierarchical_merge_tree_model}, we describe a modification to the hierarchical merge tree model with iterative segmentation accumulation. Experimental results are shown in Section~\ref{sec:experiments}, in which we compare the segmentation performance of our method with different settings, as well as with other recent state-of-the-art methods. In Section~\ref{sec:conclusion}, we summarize our current work and discuss the possible improvement for the future.

\section{Previous work}\label{sec:previous_work}
There are two different perspectives of image segmentation~\cite{arbelaez2011contour}. One is edge detection, which aims at finding edges between different perceptual pixel groups. The other one is region segmentation, which partitions an image into disjoint regions. Usually, edge detection focuses on assigning a binary label to each pixel with certain confidence indicating if it belongs to an edge or not and does not guarantee closed object contours. Though closed contours and thus regions they encircle can be recovered from edges, such transformation with high accuracy is usually non-trivial. On the other hand, region segmentation seeks to find the cluster membership of each pixel, and closed contours of an object can be trivially generated as the outmost points of a region. Many region segmentation methods also take advantage of the edge detection outputs as boundary cues to help with the search for correct partitioning. Our method belongs to the region segmentation category, and in this section we emphasize reviewing previous related works in this category.

First, we briefly summarize related edge detection works. Early edge detections are mostly based on image derivatives~\cite{marr1980theory,canny1986computational} or filter banks responses~\cite{morrone1987feature,freeman1991design}. More recent works utilize richer information such as colors and textures. One of the most notable works, gPb~\cite{arbelaez2011contour}, combines multi-scale local cues and globalized cues via spectral clustering and sets up a benchmark for edge detection and region segmentation research. Taking advantage of supervised learning techniques has also become the recent trend in edge detection. Ren and Bo~\cite{ren2012discriminatively} train a classifier with sparse codes on local neighborhood information and improve the edge detection performance. Dollar and Zitnick~\cite{dollar2013structured} propose a structured learning framework using modified random decision forest for efficient edge detection. Seyedhosseini and Tasdizen~\cite{seyedhosseini2015semantic} propose a hierarchical model to capture multi-scale contextual information and achieve state-of-the-art edge detection performance.

Early works on region segmentation seek to directly group image pixels in an unsupervised manner. Belongie \emph{et al.}~\cite{belongie1998color} fit Gaussian mixture models to cluster pixels based on six-dimensional color and texture features. Mean shift~\cite{comaniciu2002mean} and its variant~\cite{vedaldi2008quick} consider region segmentation as a density mode searching problem. A number of works belong to graph partitioning category, which regards an image as a graph with pixels being nodes and edge weights indicating dissimilarity between neighbor pixels. Normalized cuts~\cite{shi2000normalized} takes the image affinity matrix and partitions an image by solving eigenvalue problems. Felzenszwalb and Huttenlocher~\cite{felzenszwalb2004efficient} propose to greedily merge two connected components if there exists an inter-component edge weight that is less than the largest edge weights in the minimum spanning trees of both components. Arbelaez \emph{et al.}~\cite{arbelaez2011contour} propose a variant of watershed transform to generate a hierarchy of closed contours. We refer readers to~\cite{zhu2016beyond} for a comprehensive review of existing methods.

As in edge detection, supervised learning based methods for region segmentation have gained increased popularity in recent years. This trend leads to and is further promoted by a number of publicly available computer vision data sets with human-labeled ground truth~\cite{martin2001database,arbelaez2011contour,shotton2006textonboost,gould2009decomposing,voc2012,silberman2012indoor}. Though unsupervised methods, such as~\cite{cheng2011multi,zhu2013object}, are shown to generate perceptually coherent segmentations, learning segmentation models from supervised data enables much more capability and flexibility of incorporating preference from human observers and leads to many more interesting works.

Following the classic foreground/background segmentation, object-independent segmentation methods seek to partition an image based only on its appearance and do not utilize underlying semantics about the scene or specific information about target objects. Kim \emph{et al.} propose a hypergraph-based correlation clustering framework~\cite{kim2014image} that uses structured SVM for learning the structural information from training data. Arbelaez \emph{et al.} develop the multi-scale combinatorial grouping (MCG) framework~\cite{arbelaez2014multiscale} that exploits multi-scale information and uses a fast normalized cuts algorithm for region segmentation. Yu \emph{et al.}~\cite{yu2015piecewise} present a piecewise flat embedding learning algorithm and report the best published results so far on Berkeley Segmentation Data Set using the MCG framework. Two other recent superpixel-merging approaches are ISCRA~\cite{ren2013image} and GALA~\cite{nunez2013machine}. Starting with a fine superpixel over-segmentation, ISCRA adaptively divides the whole region merging process into different cascaded stages and trains a respective logistic regression model at each stage to determine the greedy merging. Meanwhile, GALA improves the boundary classifier training by augmenting the training set via repeatedly iterating through the merging process. Moreover, impressive results in the extensive evaluations on six public segmentation data sets are reported in~\cite{ren2013image}.

Object-dependent or semantic segmentation is another branch of region segmentation. Object-dependent prior knowledge is exploited to guide or improve the segmentation process. Borenstein and Ullman~\cite{borenstein2008combined} formulate object segmentation as a joint model that uses both low-level visual cues and high-level object class information. Some other object segmentation methods first generate object segmentation hypotheses using low-/mid-level features and then rank segments with high-level prior knowledge~\cite{carreira2012cpmc,arbelaez2012semantic}. A recent work, SCALPEL~\cite{weiss2013scalpel}, incorporates high-level information in the segmentation process and can generate object proposals more efficiently and accurately. There are also a group of methods, called co-segmentation, that utilizes the homogeneity between different target objects and jointly segments multiple images simultaneously~\cite{joulin2010discriminative,vicente2011object,kim2011distributed}.

Our method falls into the object-independent hierarchical segmentation category. A preliminary version of our method with the merge tree model and a greedy inference algorithm appeared in~\cite{liu2012watershed,liu2014modular} and was only applied to segmenting electron microscopy images, apart from which the contributions of this paper include:
\begin{itemize}
\item Reformulation of the hierarchical merge tree as a constrained conditional model with globally optimal solutions defined and an efficient inference algorithm developed, instead of the greedy tree model in~\cite{liu2012watershed,liu2014modular}.
\item An iterative approach to diversify merge tree generation and improve results via segmentation accumulation.
\item Experiments that extensively compare different variants and settings of the hierarchical merge tree model and show the robustness of the proposed approach against image noise at testing time.
\item Experiments with state-of-the-art results on six public data sets for general image segmentation.
\end{itemize}

Compared with recent competitive hierarchical segmentation methods, ISCRA~\cite{ren2013image} and GALA~\cite{nunez2013machine}, which use a threshold-based greedy region merging strategy, our hierarchical merge tree model has two major advantages. First, the tree structure enables the incorporation of higher order image information into segmentation. The merge/split decisions are made together in a globally optimal manner instead of by looking only at local region pairs. Second, our method does not require the threshold parameter to determine when to stop merging as in ISCRA and GALA, which may be so important to the results that needs carefully tuning. Furthermore, our method is almost parameter-free given the initial superpixel over-segmentation. The only parameter is the number of iterations, which can be fixed as shown in the experiments on all the data sets.

\section{Hierarchical merge tree model}\label{sec:hierarchical_merge_tree_model}
Given an image $I$ consisting of pixels $\mathcal{P}$, a segmentation is a partition of $\mathcal{P}$, denoted as $S=\{s_i\in 2^{\mathcal{P}}\,|\,\cup_is_i=\mathcal{P};\forall i\neq j,s_i\cap s_j=\varnothing\}$, where $2^{\mathcal{P}}$ is the power set of $\mathcal{P}$. A segmentation assigns every pixel an integer label that is unique for each image object. Each $s_i$, which is a connected subset of pixels in $\mathcal{P}$, is called a segment or region. All possible partitions form a segmentation space $\mathcal{S}_{\mathcal{P}}$. A ground truth segmentation $S_g\in\mathcal{S}_{\mathcal{P}}$ is usually generated by humans and considered as the gold standard. The accuracy of a segmentation $S$ is measured based on its agreement with $S_g$. In a probabilistic setting, solving a segmentation problem is formulated as finding a segmentation that maximizes its posterior probability given the image as
\begin{equation}
S^*=\underset{S\in\mathcal{S}_{\mathcal{P}}}{\arg\max}\ P(S\,|\,I).\label{eq:seg_post_prob}
\end{equation}

The current trend to alleviate the difficulty in pixelwise search for $S^*$ is to start with a set of over-segmenting superpixels. A superpixel is an image segment consisting of pixels that have similar visual characteristics. A number of algorithms~\cite{shi2000normalized,felzenszwalb2004efficient,levinshtein2009turbopixels,veksler2010superpixels,achanta2012slic} can be used to generate superpixels. In this paper, we use the watershed algorithm~\cite{beucher1979use} over the output of the boundary detector gPb~\cite{arbelaez2011contour}.

Let $S_o$ be the initial over-segmentation given by the superpixels, the final segmentation consisting only of merged superpixels in $S_o$ can be represented as $S=\{s_i\in2^{\mathcal{P}}\,|\,\cup_is_i=\mathcal{P};\forall i\neq j,s_i\cap s_j=\varnothing;\forall i,\exists S'\in2^{S_o},\textrm{s.t. }s_i=\cup_{s'_j\in S'}s'_j\}$. Therefore, the search space for $S$ is largely reduced to $\mathcal{S}\subseteq\mathcal{S}_{\mathcal{P}}$. Even so, however, exhaustive search is still intractable, and some kind of heuristic has to be injected. We propose to further limit $\mathcal{S}$ to a set of segmentations induced by tree structures and make the optimum search feasible.

\subsection{Hierarchical merge tree}\label{sec:hierarchical_merge_tree}
\begin{figure*}[!t]
  \centering
  \subfloat[\label{subfig:toy_segi}]{\includegraphics[width=0.2\linewidth]{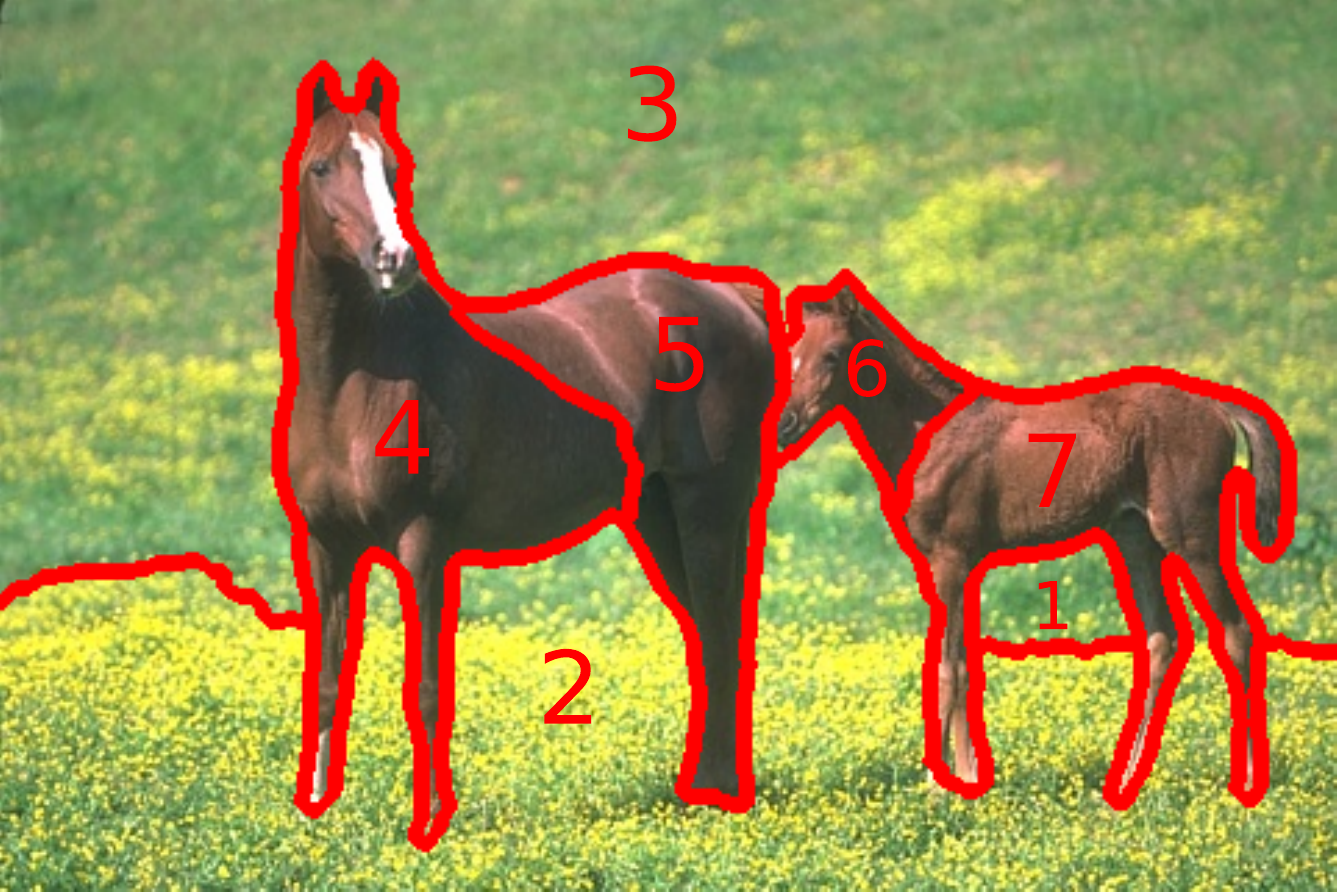}}\hfill
  \subfloat[\label{subfig:toy_seg}]{\includegraphics[width=0.2\linewidth]{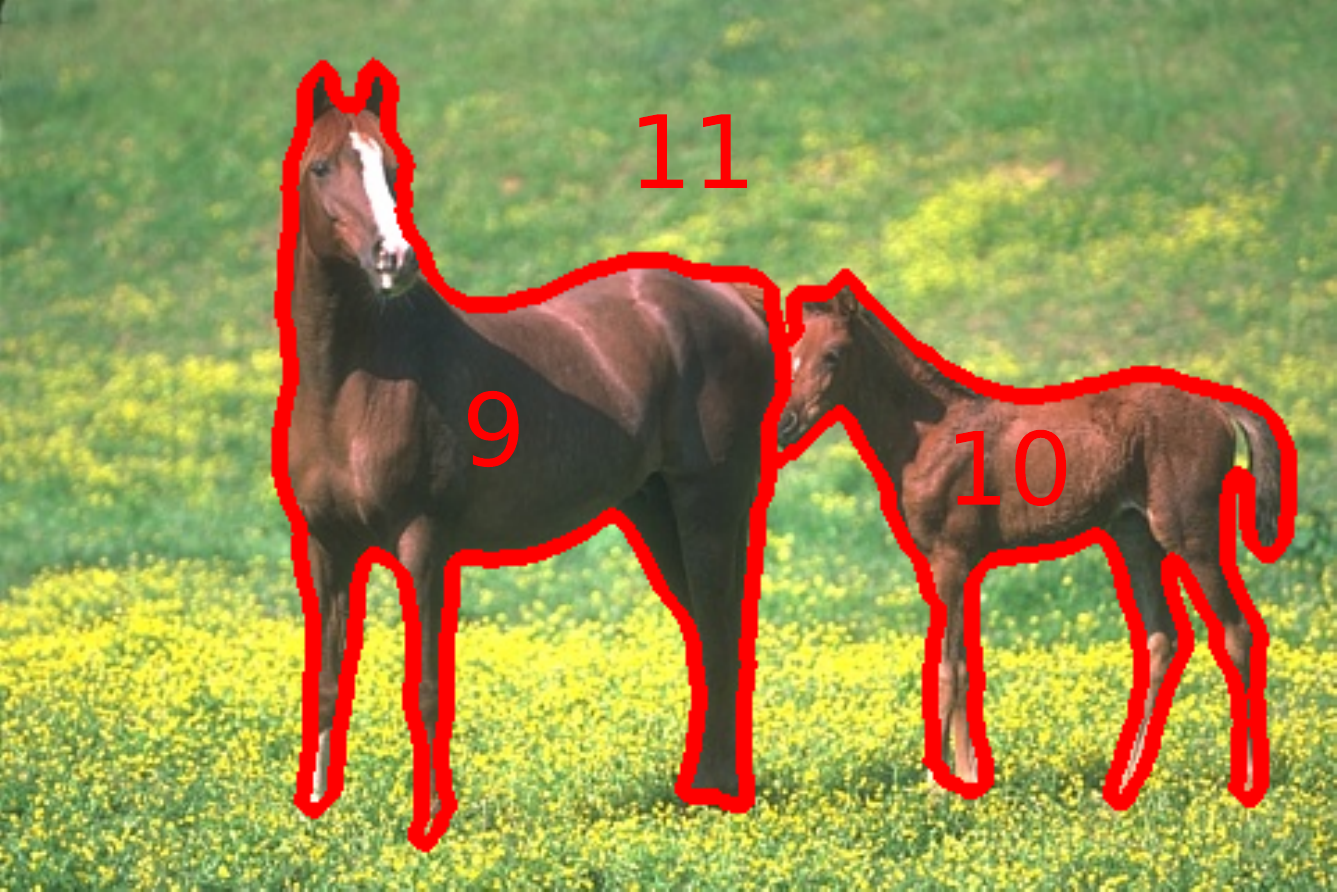}}\hfill
  \subfloat[\label{subfig:toy_tree}]{\includegraphics[width=0.28\linewidth]{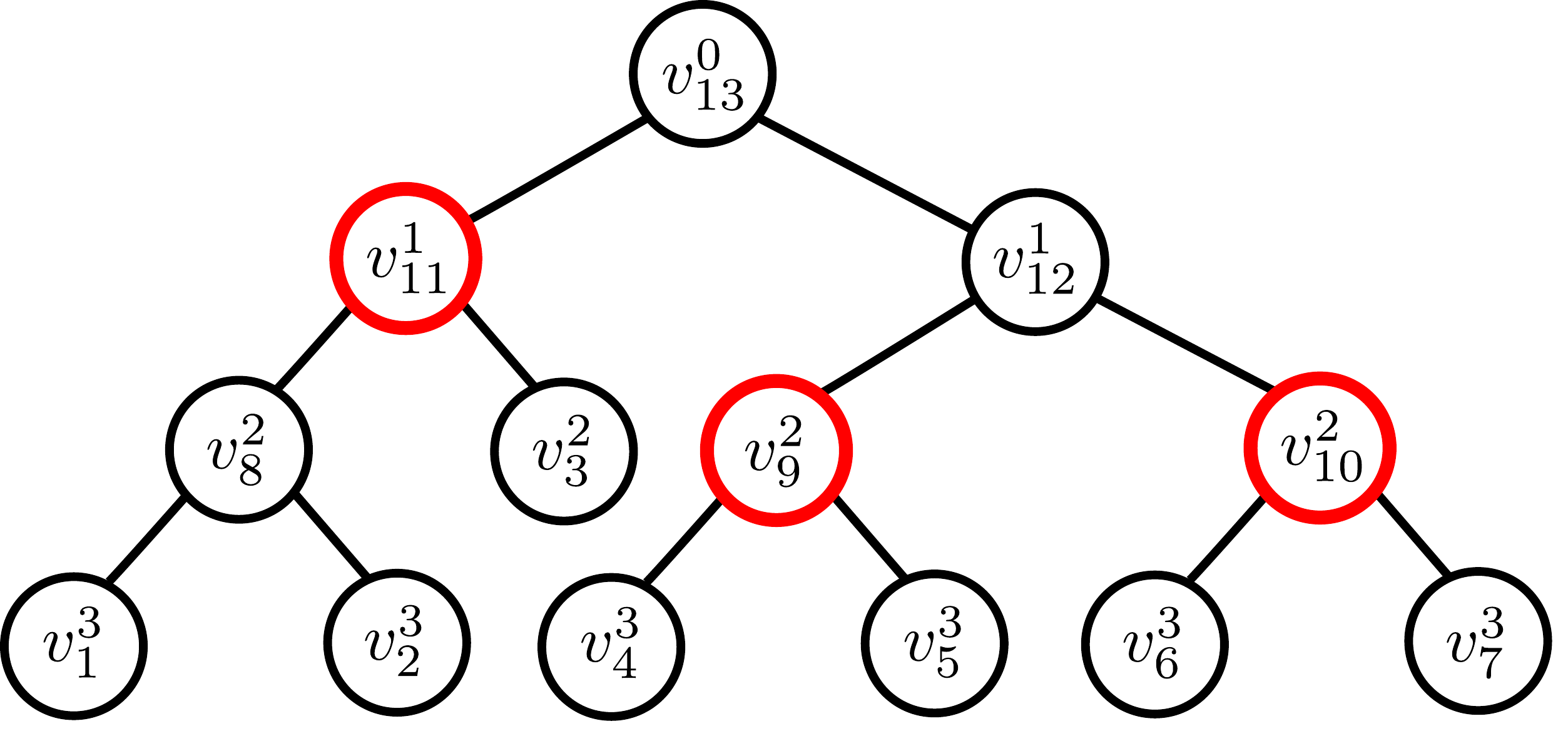}}\hfill
  \subfloat[\label{subfig:toy_tree_ccm}]{\includegraphics[width=0.28\linewidth]{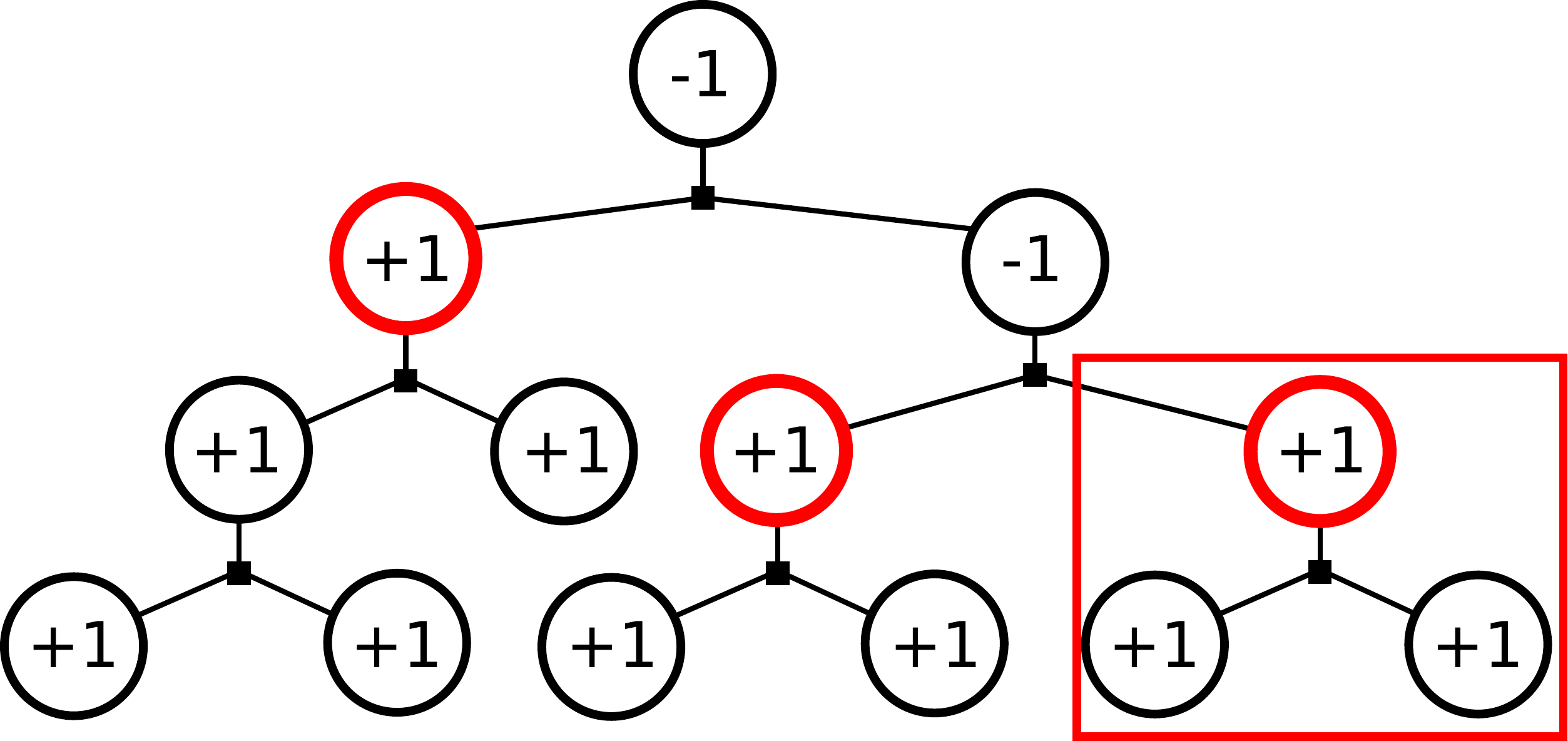}}
  \caption{Example of (a) an initial segmentation, (b) a consistent final segmentation, (c) a merge tree, and (d) the corresponding conditional model factor graph (Section~\ref{sec:ccm}) with correct labeling. In (c), the leaf nodes have labels identical to those of the initial regions. The red nodes correspond to regions in the final segmentation. The red box in (d) indicates a clique in the model.}
  \label{fig:toy_segi_seg_tree}
\end{figure*}

Consider a graph, in which each node corresponds to a superpixel, and an edge is defined between two nodes that share boundary pixels with each other. Starting with the initial over-segmentation $S_o$, finding a final segmentation, which is essentially the merging of initial superpixels, can be considered as combining nodes and removing edges between them. This superpixel merging can be done in an iterative fashion: each time a pair of neighboring nodes are combined in the graph, and corresponding edges are updated. To represent the order of such merging, we use a full binary tree structure, which we call the hierarchical merge tree (or merge tree for short) throughout this paper. In a merge tree $Tr=(\mathcal{V},\mathcal{E})$, a node $v^d_i\in\mathcal{V}$ represents an image segment $s_i\in2^{\mathcal{P}}$, where $d$ denotes the depth in $Tr$ at which this node occurs. Leaf nodes correspond to initial superpixels in $S_o$. A non-leaf node corresponds to an image region formed by merging superpixels, and the root node corresponds to the whole image as one single region. An edge $e_{ij}\in\mathcal{E}$ between node $v^d_i$ and its child $v^{d+1}_j$ exists when $s_j\subset s_i$, and a local structure $(\{v^d_i,v^{d+1}_j,v^{d+1}_k\},\{e_{ij},e_{ik}\})$ represents $s_i=s_j\cup s_k$. In this way, finding a final segmentation becomes finding a subset of nodes in $Tr$. Fig.~\ref{subfig:toy_tree} shows a merge tree example with initial superpixels shown in Fig.~\ref{subfig:toy_segi} corresponding to the leaf nodes. The non-leaf nodes represent image regions as combinations of initial superpixels. Fig.~\ref{subfig:toy_seg} shows a final segmentation formed by a subset of tree nodes. It is noteworthy that a merge tree defined here can be seen as a dendrogram in hierarchical clustering~\cite{duda1973pattern} with each cluster being an image region.

In order to determine the merging priority, we define a merging saliency function $f_{ms}:S^2\rightarrow\mathbb{R}$ that assigns a real number to each pair of regions in $S$ as a measurement of their merging likelihood. For any pair of regions $s_i$ and $s_j$ that are not neighbors, we define $f_{ms}(s_i,s_j)=-\infty$. Then starting from a set of initial superpixels $S=S_o$ as leaf nodes, a merge tree is constructed by iteratively merging $(s^*_i,s^*_j)=\arg\max_{s_i,s_j\in S,i\neq j}f_{ms}(s_i,s_j)$ to a parent node, until only one region remains in $S$ corresponding to the root node. Statistics over the strengths of boundary pixels between two merging regions from boundary detection probability maps may be used as $f_{ms}$. Following~\cite{liu2014modular}, we use negated median
\begin{equation}
f_{ms}(s_i,s_j)=1-\textrm{median}(\left\{Pb(k)\,|\,k\in\mathcal{B}(s_i,s_j)\right\}),
\end{equation}
where $Pb(k)$ is the value of the $k$-th pixel on some boundary detection probability map $Pb$, and $\mathcal{B}(s_i,s_j)$ is the set of boundary pixels between $s_i$ and $s_j$. $\mathcal{B}$ can be different on implementation. In our work, we define
\begin{equation}
\mathcal{B}(s_i,s_j)=\left(s_i\cap\mathcal{N}(s_j)\right)\cup\left(s_j\cap\mathcal{N}(s_i)\right),\label{eq:boundary_pixels}
\end{equation}
where $\mathcal{N}(s_{\cdot})$ is the set of neighbor pixels of $s_{\cdot}$. We also propose to learn the merging saliency function from data in Section~\ref{sec:iterative_hierarchical_merge_tree_model}.

\subsection{Constrained conditional model}\label{sec:ccm}
In order to select a subset of nodes that forms an optimal segmentation, we formulate the merge tree as a constrained conditional model. It is essentially a factor graph for the merge tree, in which the node set aligns identically with $\mathcal{V}$, and each merge in the merge tree that involves three nodes $(\{v^d_i,v^{d+1}_j,v^{d+1}_k\},\{e_{ij},e_{ik}\})$ is considered as a clique $p_i$ in the factor graph. A label $y_i=+1$ or $y_i=-1$ is assigned to each node indicating whether its children merge or not. All leaf nodes must be labeled $+1$. A complete label assignment $Y=\{y_i\}_i$ of all nodes must also be subject to the region consistency constraint that if a node is labeled $+1$, all of its descendants must be labeled $+1$ as well. Then the nodes whose labels are $+1$ and parents are labeled $-1$ are selected as segments in the final segmentation. Fig.~\ref{subfig:toy_tree_ccm} is the factor graph for the constrained conditional model derived from the merge tree in Fig.~\ref{subfig:toy_tree}. The red box shows a clique, and a set of consistent labeling is shown.

We train a classifier (Section~\ref{sec:bc}) to predict the probability $P(y_i)$ for each merge $(\{v^d_i,v^{d+1}_j,v^{d+1}_k\},\{e_{ij},e_{ik}\})$. Then we score each clique $p_i$ by associating it with energy with respect to its label
\begin{equation}
E_i(y_i)=-\log P(y_i),y_i=\pm1.
\end{equation}

Under the Markov assumption, we formulate our labeling problem as a constrained optimization problem
\begin{equation}
\begin{aligned}
& \underset{Y}{\text{min}} & & \sum_{y_i\in Y}E_i(y_i),y_i=\pm1,\\
& \text{s.t.} & & y_i=+1,\forall i,v^d_i\text{ is a leaf},\\
& & & y_i\leq y_j,\forall i,j,v^d_i\text{ is parent to }v^{d+1}_j,
\end{aligned}\label{eq:energy_opt}
\end{equation}
for which an inference algorithm will be introduced in Section~\ref{sec:inf}.

\subsection{Boundary classifier}\label{sec:bc}
To score each clique, we train a boundary classifier to predict the probability of each merge. To generate training labels that indicate whether the boundary between two regions exists or not, we compare both the merge and the split case against the ground truth under certain error metric, such as the Rand error~\cite{rand1971objective} and the variation of information (VI)~\cite{meila2005comparing,yang2008unsupervised} (See Section~\ref{sec:eval_metrics} for details). The case with smaller error deviates less from the ground truth and is adopted. In practice, we choose VI for its robustness to size rescaling~\cite{nunez2013machine}.

Boundary features and region features are extracted for classification. For a pair of merging regions, boundary features provide direct cues about how it is likely the boundary truly exists, and regional features measure geometric and textural similarities between the two regions, which can both be informative to boundary classification. We choose features following~\cite{ren2013image} for comparison purposes. A summary of features is provided in Appendix~\ref{app:feat}. The boundary classifier is not limited to any specific supervised classification model. We use random forest~\cite{breiman2001random} in our experiments. The parameter setting for our random forest is summarized in Appendix~\ref{app:param}.

The boundary classification problem is highly non-linear, and learning one universally good boundary classifier for all merging cases is essentially difficult. The size of merging regions affects the feature representativeness in classification. For instance, textural features in the form of averaged histograms among patches may not be informative when the merging regions are too small, because textural features can be extracted from only a very limited number of image patches and is thus noisy. On the other hand, when two regions are so big that they contain under-segmentation from different perceptual groups, the features again may not be meaningful, but for a different reason, that is, the histogram averaging is not able to represent the variation of textures. It is worth noting that for the same reason, different classifiers have to be learned at different merging stages in~\cite{ren2013image}.

We categorize the classification problem into sub-problems, train a separate sub-classifier for each sub-problem, and form the boundary classifier as an ensemble of sub-classifiers. We compute the median size $|s|_{\textrm{med}}$ of all regions observed in the training set and assign a category label to a training sample that involves regions $s_i$ and $s_j$ based on their sizes as in~\eqref{eq:boundary_classifier_category}. Three sub-classifiers are then trained respectively using only samples with identical category labels.
\begin{equation}\label{eq:boundary_classifier_category}
c(s_i,s_j)=\left\{
\begin{array}{l}
1\textrm{, if }\max(|s_i|,|s_j|)<|s|_{\textrm{med}},\\
2\textrm{, if }\min(|s_i|,|s_j|)<|s|_{\textrm{med}}\leq\max(|s_i|,|s_j|),\\
3\textrm{, otherwise.}
\end{array}
\right.
\end{equation}

At testing time, a sample is categorized based on its region sizes and assigned to the corresponding sub-classifier for prediction. Since all the sub-classifiers are always used adjointly, we refer to the set of all sub-classifiers as the boundary classifier in the rest of this paper.

\subsection{Inference}\label{sec:inf}
Exhaustive search to solve~\eqref{eq:energy_opt} has exponential complexity. Given the tree structure, however, we can use a bottom-up/top-down algorithm to efficiently find the exact optimal solution under the region consistency constraint. The fundamental idea of the bottom-up/top-down algorithm is dynamic programming: in the bottom-up step, the minimum energies for both decisions (merge/split) under the constraint are kept and propagated from leaves to the root, based on which the set of best consistent decisions is made from the root to leaves in the top-down step. It is noteworthy that our bottom-up/top-down algorithm is only for inference and conceptually different from the top-down/bottom-up framework in~\cite{borenstein2008combined}, which seeks to combine high-level semantic information and low-level image features. On the other hand, the two-way message passing algorithm used in~\cite{borenstein2008combined} and our algorithm both belong to the Pearl's belief propagation~\cite{pearl1982reverend,kschischang2001factor} category, except that our inference algorithm explicitly incorporates the consistency constraint into the optimization procedure.

In the bottom-up step, a pair of energy sums are kept track of for each node $v^d_i$ with children $v^{d+1}_j$ and $v^{d+1}_k$: the merging energy $E^m_i$ of node $v^d_i$ and its descendants all being labeled $+1$ (merge), the splitting energy $E^s_i$ of it that $v^d_i$ is labeled $-1$ (split), and its descendants are labeled optimally subject to the constraint. Then the energies can be computed bottom-up recursively as
\begin{align}
&E^m_i=E^m_j+E^m_k+E_i(y_i=+1),\\
&E^s_i=\min(E^m_j,E^s_j)+\min(E^m_k,E^s_k)+E_i(y_i=-1).
\end{align}
For leaf nodes, we assign $E^m_i=0$ and $E^s_i=\infty$ to enforce their being labeled $+1$. Fig.~\ref{alg:bottom_up} illustrates the bottom-up algorithm in pseudocode.


\begin{figure}[!t]
\begin{algorithmic}[1]
\Require A list of energy $\{E_i(y_i)\}_{i=1}^{|\mathcal{V}|}$ for each clique $p_i$
\Ensure A list of energy tuples $\mathcal{T}_E=\{(E^m_i,E^s_i)\}_{i=1}^{|\mathcal{V}|}$
\State $\mathcal{T}_E\gets\{\}$
\State $\textbf{ComputeEnergyTuples}(v^0_r)$, where $v^0_r$ is the root node
\State /* \emph{Helper function that recursively computes energies} */
\Function{\textbf{ComputeEnergyTuples}}{$v^d_i$}:
\If{$v^d_i$ is a leaf node}
\State $\mathcal{T}_E\gets\mathcal{T}_E\cup\{(0,\infty)\}$
\State \textbf{return} $(0,\infty)$
\EndIf
\State $(E^m_j,E^s_j)\gets\textrm{\textbf{ComputeEnergyTuples}}(v^{d+1}_j)$\label{ln:bu_0}
\State $(E^m_k,E^s_k)\gets\textrm{\textbf{ComputeEnergyTuples}}(v^{d+1}_k)$
\State $E^m_i\gets E^m_j+E^m_k+E_i(y_i=+1)$
\State \parbox[t]{\dimexpr\linewidth-\algorithmicindent}{$E^s_i\gets\min(E^m_j,E^s_j)+\min(E^m_k,E^s_k)+E_i(y_i=-1)$}
\State \textbf{return} $(E^m_i,E^s_i)$\label{ln:bu_1}
\EndFunction
\end{algorithmic}
\caption{Pseudocode of the bottom-up energy computation algorithm.}\label{alg:bottom_up}
\end{figure}

In the top-down step, we start from the root and do a depth-first search: if the merging energy of a node is lower than its splitting energy, label this node and all its descendants $+1$; otherwise, label this node $-1$ and search its children. Fig.~\ref{alg:top_down} illustrates the top-down algorithm in pseudocode.

\begin{figure}[!t]
\begin{algorithmic}[1]
\Require A list of energy tuples $\mathcal{T}_E=\{(E^m_i,E^s_i)\}_{i=1}^{|\mathcal{V}|}$
\Ensure A complete label assignment $Y=\{y_i\}_{i=1}^{|\mathcal{V}|}$
\State $Y\gets\{\}$
\State $\textbf{AssignNodeLabels}(v^0_r)$, where $v^0_r$ is the root node
\State /* \emph{Helper function that recursively decides node labels} */
\Function{\textbf{AssignNodeLabels}}{$v^d_i$}:
\If{$E^m_i<E^s_i$}
\State $Y\gets Y\cup\{y_i=+1\}\cup\{y_{i'}=+1\,|\,\forall i'\in\mathcal{D}(i)\}$, where $\mathcal{D}(i)$ is set of indices of descendants of $v^d_i$
\Else
\State $Y\gets Y\cup\{y_i=-1\}$
\State $\textbf{AssignNodeLabels}(v^{d+1}_j)$
\State $\textbf{AssignNodeLabels}(v^{d+1}_k)$
\EndIf
\EndFunction
\end{algorithmic}
\caption{Pseudocode of the top-down label assignment algorithm.}\label{alg:top_down}
\end{figure}

Eventually, we select the set of the nodes, such that its label is $+1$ and its parent is labeled $-1$, to form an optimal final segmentation. In both algorithms, each node is visited exactly once with constant operations, and we need only linear space proportional to the number nodes for $\mathcal{T}_E$ and $Y$, so the time and space complexity are both $O(|\mathcal{V}|)$.

\section{Iterative hierarchical merge tree model}\label{sec:iterative_hierarchical_merge_tree_model}
The performance upper bound of the hierarchical merge tree model is determined by the quality of the tree structure. If all true segments exist as nodes in the tree, they may be picked out by the inference algorithm using predictions from well-trained boundary classifiers. However, if a desirable segment is not represented by any node in the tree, the model is not able to recover the segment. Hence, the merging saliency function, which is used to determine merging priorities, is critical to the entire performance. With a good merging saliency function, we can push the upper bound of performance and thus improve segmentation accuracy.

Statistics over the boundary strengths can be used to indicate merging saliency. We use the negated median of boundary pixel strengths as the initial representation of saliency, as mentioned in Section~\ref{sec:hierarchical_merge_tree}. Since a boundary classifier is essentially designed to measure region merging likelihood, and it has advantages over simple boundary statistics because it takes various features from both boundary and regions, we propose to use the merging probabilities predicted by boundary classifiers as the merging saliency to construct a merge tree.

\begin{figure*}[!t]
  \centering
  \subfloat[\label{subfig:toy_itr_tr}]{\includegraphics[height=456\in]{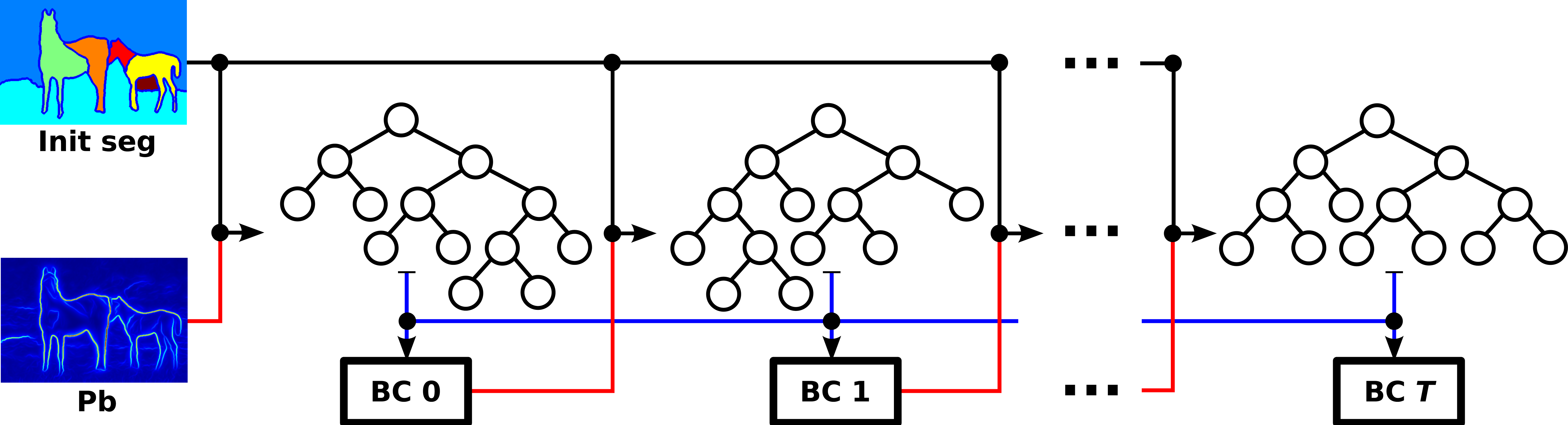}}\\
  \subfloat[\label{subfig:toy_itr_te}]{\includegraphics[height=610.5\in]{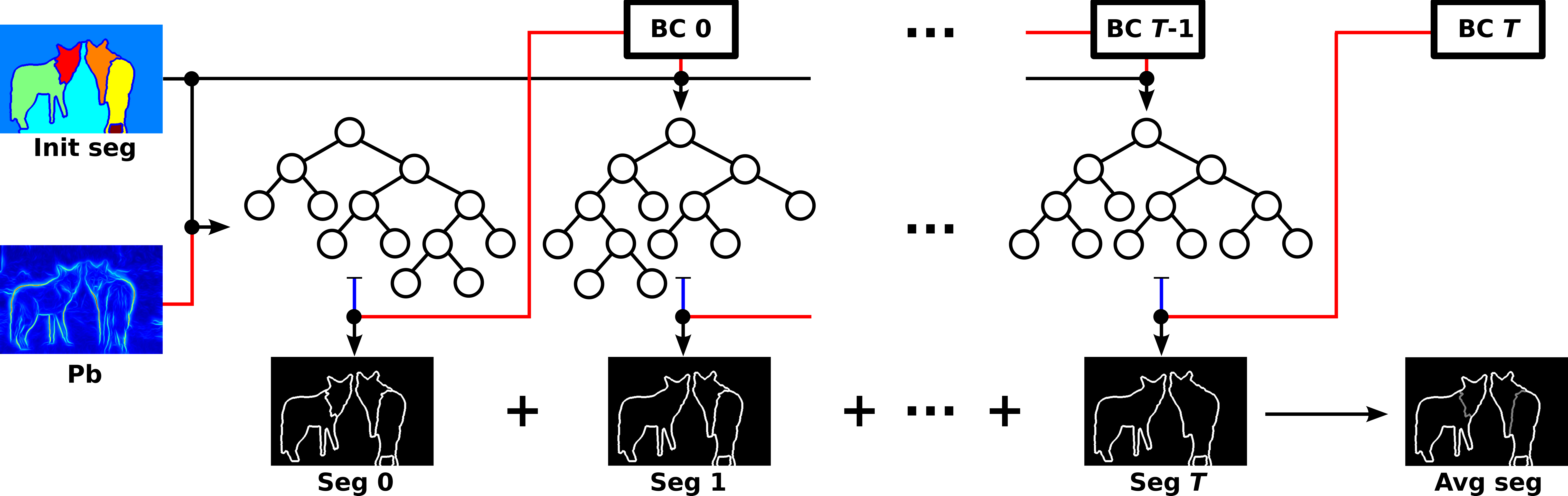}}
  \caption{Illustrations of (a) the training and (b) the testing procedure of the iterative hierarchical merge tree model. Starting with the fixed initial superpixels (``Init Seg''), the first iteration uses boundary probability (``Pb'') statistics for merge tree generation, and the training procedure iteratively augments the training set by incorporating new samples from merge trees and trains a new boundary classifier (``BC''), which is used for merge tree generation from the same initial superpixels in the next iteration. At testing time, boundary probability statistics and boundary classifiers learned at each iteration are used to generate merge trees from the same initial superpixels, and each boundary classifier is used to score merge cliques in the previous iteration; segmentations are generated from each merge tree and accumulated to generate the final contour hierarchy. The black lines show the use of initial superpixels, the red lines show the use of boundary classifiers, and the blue lines show the flow of sample data collected from tree structures.}
  \label{fig:toy_itr}
\end{figure*}

As described in Section~\ref{sec:bc}, the training of a boundary classifier requires samples generated from a merge tree, but we would like to use a boundary classifier to construct a merge tree. Therefore, we propose an iterative approach that alternately collects training samples from a merge tree for the training of boundary classifiers and constructs a merge tree with the trained classifier. As illustrated in Fig.~\ref{subfig:toy_itr_tr}, we initially use the negated median of boundary strengths to construct a merge tree, collect region merging samples, and train a boundary classifier $f_b^0$. Then, the boundary classifier $f_b^0$ is used to generate a new merge tree from the same initial superpixels $S_o$, from which new training samples are generated. We next combine the samples from the current iteration and from the previous iterations, remove duplicates, and train the next classifier $f_b^1$. This process is repeated for $T$ iterations or until the segmentation accuracy on a validation set no longer improves. In practice, we fix the iteration number to $T=10$ for all data sets. Eventually, we have a series of boundary classifiers $\{f_b^t\}_{t=0}^T$ from each training iteration. The training algorithm is illustrated in Fig.~\ref{alg:itr_tr}.

\begin{figure}[!t]
\begin{algorithmic}[1]
\Require Original images $\{I_i\}_{i=1}^{N_{\textrm{tr}}}$, boundary maps $\{Pb_i\}_{i=1}^{N_{\textrm{tr}}}$, and iteration number $T$
\Ensure Boundary classifiers $\{f^t_b\}_{t=0}^T$
\State Generate initial superpixels $\{S_{oi}\}_{i=1}^{N_{\textrm{tr}}}$
\For{$t:0,1,\ldots,T$}
\If{$t==0$}
\State Generate $\{Tr^0_i\}_{i=1}^{N_{\textrm{tr}}}$ from $\{S_{oi}\}_{i=1}^{N_{\textrm{tr}}}$ using $\{Pb_i\}_{i=1}^{N_{\textrm{tr}}}$
\Else
\State Generate $\{Tr^t_i\}_{i=1}^{N_{\textrm{tr}}}$ from $\{S_{oi}\}_{i=1}^{N_{\textrm{tr}}}$ using $f^{t-1}_b$
\EndIf
\State Generate samples $\{(X^t_i,Y^t_i)\}_{i=1}^{N_{\textrm{tr}}}$ from $\{Tr^t_i\}_{i=1}^{N_{\textrm{tr}}}$
\State Train $f^t_b$ using $\cup_{t'=1}^t\{(X^{t'}_i,Y^{t'}_i)\}_{i=1}^{N_{\textrm{tr}}}$
\EndFor
\end{algorithmic}
\caption{Pseudocode of the iterative training algorithm.}\label{alg:itr_tr}
\end{figure}

At testing time, we take the series of trained classifiers and iterate in a way similar to the training process, as shown in Fig.~\ref{subfig:toy_itr_te}: at each iteration $t$, we take the previous boundary classifier $f_b^{t-1}$ to construct a merge tree over the same initial superpixels $S_o$ and use the current classifier $f_b^t$ to predict each merge score in the merge tree, based on which a final segmentation $S^t$ is inferred. Finally, we transform each segmentation into a binary closed contour map by assigning boundary pixels $1$ and others $0$ and average them for each image over all iterations to generate a segmentation hierarchy in the form a real-valued contour map. The testing algorithm is illustrated in Fig.~\ref{alg:itr_te}.

\begin{figure}[!t]
\begin{algorithmic}[1]
\Require Original images $\{I_i\}_{i=1}^{N_{\textrm{te}}}$, boundary maps $\{Pb_i\}_{i=1}^{N_{\textrm{te}}}$, and boundary classifiers $\{f^t_b\}_{t=0}^T$
\Ensure Hierarchical segmentation contour map $\{C_i\}_{i=1}^{N_{\textrm{te}}}$
\State Generate initial superpixels $\{S_{oi}\}_{i=1}^{N_{\textrm{te}}}$
\For{$t:0,1,\ldots,T$}
\If{$t==0$}
\State Generate $\{Tr^0_i\}_{i=1}^{N_{\textrm{te}}}$ from $\{S_{oi}\}_{i=1}^{N_{\textrm{te}}}$ using $\{Pb_i\}_{i=1}^{N_{\textrm{te}}}$
\Else
\State Generate $\{Tr^t_i\}_{i=1}^{N_{\textrm{te}}}$ from $\{S_{oi}\}_{i=1}^{N_{\textrm{te}}}$ using $f^{t-1}_b$
\EndIf
\State Score merges with $f^t_b$ and infer segmentations $\{S^t_i\}_{i=1}^{N_{\textrm{te}}}$
\State Binarize $\{S^t_i\}_{i=1}^{N_{\textrm{te}}}$ to contour maps $\{C^t_i\}_{i=1}^{N_{\textrm{te}}}$
\EndFor
\State $\{C_i\}_{i=1}^{N_{\textrm{te}}}=\{\sum_{t=0}^TC^t_i/(T+1)\}_{i=1}^{N_{\textrm{te}}}$
\end{algorithmic}
\caption{Pseudocode of the iterative testing algorithm.}\label{alg:itr_te}
\end{figure}

The explanation for the iterative approach is two-fold. First, by collecting samples that were not seen in previous iterations, we can explore the merge sample space and in turn explore the space of merge trees generated by the classifiers trained using the augmented sample set towards the ``correct'' merge tree. Second, like a bagging algorithm, segmentation averaging through iterations tends to emphasize accurate boundaries by phasing out non-systematic errors due to incorrect tree structures or classifier mispredictions. The segmentation accumulation alleviates the difficulty of training one accurate classifier to generate segmentations by improving via averaging.

\section{Experiments}\label{sec:experiments}
We conduct experiments with two validation goals. First, we evaluate the performance of our hierarchical merge tree model with different combinations of settings. Second, we compare our method with other state-of-the-art methods. Our source code is available at \url{https://github.com/tingliu/glia}.

\subsection{Setting}\label{sec:exp_setting}
We experiment with six publicly available data sets for image segmentation:
\begin{enumerate}
\item Berkeley Segmentation Data Set 300 (BSDS300)~\cite{martin2001database}: 200 training and 100 testing natural images of size $481\times321$ pixels. Multiple ground truth segmentations are provided with different labeling of details.
\item Berkeley Segmentation Data Set 500 (BSDS500)~\cite{arbelaez2011contour}: an extension of BSDS300 with 200 new testing images of the same size, with multiple ground truth segmentations for each image.
\item MSRC Object Recognition Data Set (MSRC)~\cite{shotton2006textonboost}: 591 $320\times213$ natural images with one ground truth per image. A cleaned-up version~\cite{malisiewicz2007improving} is used, in which ``void'' regions are removed, and disconnected regions that belong to the same object class are assigned different labels in a single image.
\item PASCAL Visual Object Classes Data Set (VOC12)~\cite{voc2012}: 1449 validation images with one ground truth per image for PASCAL VOC 2012 Challenge. The average image size is $496\times360$. We use the ground truth for object segmentation and treat the object boundary pixels as background.
\item Stanford Background Data Set (SBD)~\cite{gould2009decomposing}: 715 approximately $320\times240$ images of outdoor scenes with one ground truth per image.
\item NYU Depth Data Set v2 (NYU)~\cite{silberman2012indoor}: 1449 indoor scene images with one ground truth per image. Down-sampled versions ($320\times240$)~\cite{ren2012discriminatively} are used with frame pixels cropped. Only RGB channels are used in our experiment; the depth maps are not used.
\end{enumerate}

In order to compare with the other state-of-the-art methods, we follow~\cite{ren2013image} and train our boundary classifiers with the 200 training images in BSDS300. Five ground truth segmentations are selected for each image in the order of increasing details as indicated by the number of true segments. The training and the testing are done for each detail level, and the results are combined into a segmentation hierarchy. In our performance evaluation of different configurations of the merge tree model, we test on the testing images in BSDS500. For comparisons with other methods, we test on all six data sets.

Appendix~\ref{app:feat} summarizes the features used for boundary classification, most of which follows~\cite{ren2013image}. Appendix~\ref{app:param} provides the parameters that we use in our hierarchical merge tree model experiments.

\subsection{Evaluation metrics}\label{sec:eval_metrics}
Following~\cite{arbelaez2011contour}, we use the segmentation covering~\cite{voc2012}, the probabilistic Rand index~\cite{unnikrishnan2007toward}, and the variation of information~\cite{meila2005comparing,yang2008unsupervised} for segmentation accuracy evaluation. Here, we summarize the three evaluation metrics. For more details, please refer to~\cite{arbelaez2011contour}.

The segmentation covering measures averaged matching between proposed segments with a ground truth labeling, defined as
\begin{equation}
SC(S,S_g)=\sum_{s_i\in S}\frac{|s_i|}{|\mathcal{P}|}\max_{s_{j}\in S_g}\frac{|s_i\cap s_j|}{|s_i\cup s_j|},
\end{equation}
where $\mathcal{P}$ is the set of all pixels in an image. It matches each proposed segment to a true segment, with which the proposed segment has the largest overlapping ratio, and computes the sum of such optimal overlapping ratios weighted by relative segment sizes.

The Rand index, originally proposed in~\cite{rand1971objective}, measures pairwise similarity between two multi-label clusterings. It is defined as the ratio of the number of pixel pairs that have identical labels in $S$ and $S_g$ or have different labels in $S$ and $S_g$, over the number of all pixel pairs.
\begin{equation}
RI(S,S_g)=\frac{1}{{|\mathcal{P}|\choose2}}\sum_{i<j}\mathbb{I}\left(S(i)=S(j)\wedge S_g(i)=S_g(j)\right),
\end{equation}
where $S(i)$ is the label of the $i$th pixel in $S$, and $\mathbb{I}(\cdot)$ is an indicator function that returns $1$ if the input condition is met or $0$ otherwise. The Rand error is sometimes used to refer $1-RI$. The probabilistic Rand index is the Rand index averaged over multiple ground truth labelings if available.

The variation of information measures the relative entropy between a proposed segmentation and a ground truth labeling, defined as
\begin{equation}
VI(S,S_g)=H(S\,|\,S_g)+H(S_g\,|\,S),
\end{equation}
where $H(S\,|\,S_g)$ and $H(S_g\,|\,S)$ are conditional image entropies. Denote the set of all labels in $S$ as $\mathcal{L}_S$ and the set of all labels in $S_g$ as $\mathcal{L}_{S_g}$, we have
\begin{align}
H(S\,|\,S_g)=\sum_{\substack{l\in\mathcal{L}_S,\\l_g\in\mathcal{L}_{S_g}}}P(l,l_g)\log\frac{P(l_g)}{P(l,l_g)},\label{eq:cond_entropy}
\end{align}
where $P(l_g)$ is the probability that a pixel in $S_g$ receives label $l_g$, and $P(l,l_g)$ is the joint probability that a pixel receives label $l$ in $S$ and label $l_g$ in $S_g$. $H(S_g\,|\,S)$ can be defined similarly by switching $S$ and $S_g$ in~\eqref{eq:cond_entropy}.

For each data set, segmentation results are evaluated at a universal fixed scale (ODS) for the entire data set and at a fixed scale per testing image (OIS), following~\cite{arbelaez2011contour}. The evaluated numbers are averaged over all available ground truth labelings. As pointed out in~\cite{ren2013image}, since we focus on region segmentation, the pixelwise boundary-based evaluations for contour detection results~\cite{arbelaez2011contour} are not relevant, and we use only the region-based metrics.

\subsection{Ensemble vs. single boundary classifier and constrained conditional model vs. greedy tree model}\label{sec:exp_combo}
We evaluate the performance of using single (``SC'') or ensemble boundary classifiers (``EC'') (Section~\ref{sec:bc}) with our hierarchical merge tree model. We also compare the proposed constrained conditional model (``CCM'') formulation and greedy tree model (``Greedy'') previously proposed in~\cite{liu2012watershed,liu2014modular}. The greedy tree model shares the same hierarchical merge tree structure and scores each tree node only based on local merges the node is involved with, based on which a subset of highest-scored nodes that conform with the region consistency constraint are greedily selected. The training is done using the 200 training images in BSDS300 as described in Section~\ref{sec:exp_setting}, and we show the testing results on the 200 testing images in BSDS500 in Table~\ref{tab:exp_combo}.

\begin{table}[!t]
\caption{Results on BSDS500 using the constrained conditional model (CCM) formulation or greedy tree model (Greedy)~\cite{liu2012watershed,liu2014modular} in combination with the ensemble boundary classifier (EC) or single boundary classifier (SC). The segmentation covering (Covering), the probabilistic Rand index (PRI), and the variation of information (VI) are reported for optimal data set scale (ODS) and optimal image scale (OIS).}\label{tab:exp_combo}
\centering
\begin{tabular}{ccccccc}
\cline{2-7}
& \multicolumn{2}{|c}{Covering} & \multicolumn{2}{||c}{PRI} & \multicolumn{2}{||c|}{VI}\\
\hline
\multicolumn{1}{|c}{HMT variant} & \multicolumn{1}{|c}{ODS} & \multicolumn{1}{|c}{OIS} & \multicolumn{1}{||c}{ODS} & \multicolumn{1}{|c}{OIS} & \multicolumn{1}{||c}{ODS} & \multicolumn{1}{|c|}{OIS}\\
\hline
\multicolumn{1}{|c}{CCM+EC} & \multicolumn{1}{|c}{$0.594$} & \multicolumn{1}{|c}{$0.607$} & \multicolumn{1}{||c}{$0.804$} & \multicolumn{1}{|c}{$0.809$} & \multicolumn{1}{||c}{$1.682$} & \multicolumn{1}{|c|}{$1.556$}\\
\multicolumn{1}{|c}{CCM+SC} & \multicolumn{1}{|c}{$0.573$} & \multicolumn{1}{|c}{$0.581$} & \multicolumn{1}{||c}{$0.779$} & \multicolumn{1}{|c}{$0.781$} & \multicolumn{1}{||c}{$1.690$} & \multicolumn{1}{|c|}{$1.617$}\\
\multicolumn{1}{|c}{Greedy+EC} & \multicolumn{1}{|c}{$0.587$} & \multicolumn{1}{|c}{$0.620$} & \multicolumn{1}{||c}{$0.821$} & \multicolumn{1}{|c}{$0.834$} & \multicolumn{1}{||c}{$1.737$} & \multicolumn{1}{|c|}{$1.589$}\\
\multicolumn{1}{|c}{Greedy+SC~\cite{liu2014modular}} & \multicolumn{1}{|c}{$0.582$} & \multicolumn{1}{|c}{$0.601$} & \multicolumn{1}{||c}{$0.805$} & \multicolumn{1}{|c}{$0.812$} & \multicolumn{1}{||c}{$1.748$} & \multicolumn{1}{|c|}{$1.639$}\\
\hline
\end{tabular}
\end{table}

A comparison between the first two rows in Table~\ref{tab:exp_combo} shows that using ensemble boundary classifiers outperforms using only a single boundary classifier among all metrics, which supports our claim that the classifier ensemble is better able to capture underlying merging characteristics of regions at different size scales.

Comparing the first and the third row, we can see that CCM significantly outperforms the greedy model in terms of VI, which is preferred over the other metrics for segmentation quality evaluation~\cite{nunez2013machine}. It appears that CCM is outperformed by the greedy tree model in terms of PRI, but this is because both models are trained using the labels determined based on VI (Section~\ref{sec:bc}). We perform another experiment where both are trained using the labels determined based on the Rand index, and CCM outperforms the greedy model $0.829$ vs.\ $0.826$ in terms of ODS PRI and $0.855$ vs.\ $0.848$ in terms of OIS PRI.

The fourth row shows the results using our previous work~\cite{liu2014modular}. It is clear that the proposed constrained conditional model and ensemble boundary classifier are an improvement over our previous approach without including the iterative segmentation accumulation.

\subsection{Non-iterative vs. iterative segmentation accumulation}\label{sec:exp_iteration}
We evaluate the performance of our hierarchical merge tree model with or without iterative segmentation accumulation (Section~\ref{sec:iterative_hierarchical_merge_tree_model}). The experimental setting follows the previous experiments in Section~\ref{sec:exp_combo}. The constrained conditional model formulation and ensemble boundary classifiers are adopted. Results at each iteration are shown in Table~\ref{tab:exp_iteration}.

\begin{table}[!t]
\caption{Results on BSDS500 of hierarchical merge tree model with iterative segmentation accumulation. The segmentation covering (Covering), the probabilistic Rand index (PRI), and the variation of information (VI) are reported for optimal data set scale (ODS) and optimal image scale (OIS).}\label{tab:exp_iteration}
\centering
\begin{tabular}{ccccccc}
\cline{2-7}
& \multicolumn{2}{|c}{Covering} & \multicolumn{2}{||c}{PRI} & \multicolumn{2}{||c|}{VI}\\
\hline
\multicolumn{1}{|c}{Iteration} & \multicolumn{1}{|c}{ODS} & \multicolumn{1}{|c}{OIS} & \multicolumn{1}{||c}{ODS} & \multicolumn{1}{|c}{OIS} & \multicolumn{1}{||c}{ODS} & \multicolumn{1}{|c|}{OIS}\\
\hline
\multicolumn{1}{|c}{0} & \multicolumn{1}{|c}{$0.594$} & \multicolumn{1}{|c}{$0.607$} & \multicolumn{1}{||c}{$0.804$} & \multicolumn{1}{|c}{$0.809$} & \multicolumn{1}{||c}{$1.682$} & \multicolumn{1}{|c|}{$1.556$}\\
\multicolumn{1}{|c}{1} & \multicolumn{1}{|c}{$0.601$} & \multicolumn{1}{|c}{$0.637$} & \multicolumn{1}{||c}{$0.825$} & \multicolumn{1}{|c}{$0.841$} & \multicolumn{1}{||c}{$1.661$} & \multicolumn{1}{|c|}{$1.498$}\\
\multicolumn{1}{|c}{2} & \multicolumn{1}{|c}{$0.612$} & \multicolumn{1}{|c}{$0.654$} & \multicolumn{1}{||c}{$0.829$} & \multicolumn{1}{|c}{$0.853$} & \multicolumn{1}{||c}{$1.596$} & \multicolumn{1}{|c|}{$1.432$}\\
\multicolumn{1}{|c}{3} & \multicolumn{1}{|c}{$0.618$} & \multicolumn{1}{|c}{$0.666$} & \multicolumn{1}{||c}{$0.834$} & \multicolumn{1}{|c}{$0.860$} & \multicolumn{1}{||c}{$1.564$} & \multicolumn{1}{|c|}{$1.407$}\\
\multicolumn{1}{|c}{4} & \multicolumn{1}{|c}{$0.624$} & \multicolumn{1}{|c}{$0.671$} & \multicolumn{1}{||c}{$0.834$} & \multicolumn{1}{|c}{$0.864$} & \multicolumn{1}{||c}{$1.545$} & \multicolumn{1}{|c|}{$1.391$}\\
\multicolumn{1}{|c}{5} & \multicolumn{1}{|c}{$0.624$} & \multicolumn{1}{|c}{$0.676$} & \multicolumn{1}{||c}{$0.836$} & \multicolumn{1}{|c}{$0.865$} & \multicolumn{1}{||c}{$1.544$} & \multicolumn{1}{|c|}{$1.378$}\\
\multicolumn{1}{|c}{6} & \multicolumn{1}{|c}{$0.626$} & \multicolumn{1}{|c}{$0.678$} & \multicolumn{1}{||c}{$0.835$} & \multicolumn{1}{|c}{$0.867$} & \multicolumn{1}{||c}{$1.539$} & \multicolumn{1}{|c|}{$1.374$}\\
\multicolumn{1}{|c}{7} & \multicolumn{1}{|c}{$0.628$} & \multicolumn{1}{|c}{$0.679$} & \multicolumn{1}{||c}{$0.835$} & \multicolumn{1}{|c}{$0.868$} & \multicolumn{1}{||c}{$1.532$} & \multicolumn{1}{|c|}{$1.373$}\\
\multicolumn{1}{|c}{8} & \multicolumn{1}{|c}{$0.628$} & \multicolumn{1}{|c}{$0.679$} & \multicolumn{1}{||c}{$0.835$} & \multicolumn{1}{|c}{$0.869$} & \multicolumn{1}{||c}{$1.534$} & \multicolumn{1}{|c|}{$1.370$}\\
\multicolumn{1}{|c}{9} & \multicolumn{1}{|c}{$0.628$} & \multicolumn{1}{|c}{$0.680$} & \multicolumn{1}{||c}{$0.835$} & \multicolumn{1}{|c}{$0.869$} & \multicolumn{1}{||c}{$1.530$} & \multicolumn{1}{|c|}{$1.371$}\\
\multicolumn{1}{|c}{10} & \multicolumn{1}{|c}{$0.629$} & \multicolumn{1}{|c}{$0.679$} & \multicolumn{1}{||c}{$0.835$} & \multicolumn{1}{|c}{$0.869$} & \multicolumn{1}{||c}{$1.526$} & \multicolumn{1}{|c|}{$1.375$}\\
\hline
\end{tabular}
\end{table}

We can see that despite occasional oscillations, the results are improved through iterations. The rate of improvement slows down as more iterations are included in the averaging process. More sophisticated ways of choosing segmentations to average over can be used, such as to average segmentations only from the iterations that achieve the top accuracy on some validation set. In our experiment, since we would like to compare our method with other methods, we keep the same setting for training and testing data sets and do not use a separate validation set. We fix the iteration number to $T=10$ and only report the results from averaging all the segmentations.

We also test how the iteration influences the robustness of our method to image noise. Gaussian white noise is added to the BSDS500 testing images. We experiment with different large noise variances $\sigma_n^2=0.001$ and $\sigma_n^2=0.01$, so that the noise is clearly observable, and the input images are considerably corrupted. The previous model learned with noise-free BSDS300 training images is then used for testing. We observe significant decrease in the strength of gPb boundary detection, so we lower the initial water level to $0.005$ from $0.01$ (Appendix~\ref{app:param}) for superpixel generation. We keep all other settings identical to the previous experiment and run the iterative testing (Fig.~\ref{alg:itr_te}) for $T=10$ iterations. The results for the first iteration and the last iteration
are shown in Table~\ref{tab:exp_noise}. Comparing Table~\ref{tab:exp_noise} and the corresponding entries in Table~\ref{tab:exp_iteration}, we can see that when the input images are noisy, the performance from HMT that uses gPb boundary saliency to generate the merge trees are severely degraded. However, with the iterative approach, the HMT performance is significantly improved. This is because the iterative approach enables the use of boundary classifiers that utilize different cues for better merge tree generation than using only boundary detection saliency under the noisy setting.
In addition, the iterative segmentation accumulation stabilizes the HMT performance for noisy inputs by smoothing out non-systematic errors.
\begin{table}[!t]
  \centering
  \caption{Results on BSDS500 with Gaussian white noise with different variances $\sigma_n^2$ of hierarchical merge tree model with iterative segmentation accumulation. The training uses noise-free BSDS300 training images. The segmentation covering (Covering), the probabilistic Rand index (PRI), and the variation of information (VI) are reported for optimal data set scale (ODS) and optimal image scale (OIS).}\label{tab:exp_noise}
  \begin{tabular}{|c|c|c|c||c|c||cc|}
    \cline{3-8}
    \multicolumn{1}{c}{} & \multicolumn{1}{c}{} & \multicolumn{2}{|c}{Covering} & \multicolumn{2}{||c}{PRI} & \multicolumn{2}{||c|}{VI}\\
    \hline
    \multicolumn{1}{|c}{$\sigma_n^2$} & \multicolumn{1}{|c}{Iter.} & \multicolumn{1}{|c}{ODS} & \multicolumn{1}{|c}{OIS} & \multicolumn{1}{||c}{ODS} & \multicolumn{1}{|c}{OIS} & \multicolumn{1}{||c}{ODS} & \multicolumn{1}{|c|}{OIS}\\
    \hline
    \multicolumn{1}{|c}{\multirow{2}{*}{$0.001$}} & \multicolumn{1}{|c}{0} & \multicolumn{1}{|c}{$0.457$} & \multicolumn{1}{|c}{$0.459$} & \multicolumn{1}{||c}{$0.587$} & \multicolumn{1}{|c}{$0.587$} & \multicolumn{1}{||c}{$1.929$} & \multicolumn{1}{|c|}{$1.917$}\\
    & \multicolumn{1}{|c}{10} & \multicolumn{1}{|c}{$0.617$} & \multicolumn{1}{|c}{$0.665$} & \multicolumn{1}{||c}{$0.836$} & \multicolumn{1}{|c}{$0.860$} & \multicolumn{1}{||c}{$1.576$} & \multicolumn{1}{|c|}{$1.411$}\\
    \hline
    \multicolumn{1}{|c}{\multirow{2}{*}{$0.01$}} & \multicolumn{1}{|c}{0} & \multicolumn{1}{|c}{$0.343$} & \multicolumn{1}{|c}{$0.344$} & \multicolumn{1}{||c}{$0.394$} & \multicolumn{1}{|c}{$0.394$} & \multicolumn{1}{||c}{$2.278$} & \multicolumn{1}{|c|}{$2.271$}\\
    & \multicolumn{1}{|c}{10} & \multicolumn{1}{|c}{$0.574$} & \multicolumn{1}{|c}{$0.606$} & \multicolumn{1}{||c}{$0.805$} & \multicolumn{1}{|c}{$0.815$} & \multicolumn{1}{||c}{$1.754$} & \multicolumn{1}{|c|}{$1.581$}\\
    \hline
  \end{tabular}
\end{table}

\subsection{Comparisons with other methods}
In this section, we compare our proposed iterative hierarchical merge tree method (CCM + ensemble boundary classifier + iteration, under name ``HMT'') with various other state-of-the-art region segmentation methods and benchmarks~\cite{arbelaez2011contour,ren2013image,nunez2013machine,arbelaez2014multiscale,donoser2014discrete,kim2014image,yu2015piecewise} in very recent years on the public data sets. The results are shown in Table~\ref{tab:exp_others}. Note that~\cite{kim2014image} generates a single segmentation instead of contour hierarchies for each image. The OIS evaluations are therefore essentially the same as the ODS results, so we exclude the OIS entries for the sake of clarity. Fig.~\ref{fig:exp_others} shows sample testing segmentation results for each data set.

\begin{table}[!t]
\centering
\caption{Results of different methods on (a) BSDS300, (b) BSDS500, (c) MSRC, (d) VOC12, (e) SBD, and (f) NYU data set. The segmentation covering (Covering), the probabilistic Rand index (PRI), and the variation of information (VI) are reported for optimal data set scale (ODS) and optimal image scale (OIS).}\label{tab:exp_others}
\subfloat[\label{subtab:exp_others_bsds300}]{
\begin{tabular}{ccccccc}
\cline{2-7}
& \multicolumn{6}{|c|}{\textbf{BSDS300}}\\
\cline{2-7}
& \multicolumn{2}{|c}{Covering} & \multicolumn{2}{||c}{PRI} & \multicolumn{2}{||c|}{VI}\\
\hline
\multicolumn{1}{|c}{Method} & \multicolumn{1}{|c}{ODS} & \multicolumn{1}{|c}{OIS} & \multicolumn{1}{||c}{ODS} & \multicolumn{1}{|c}{OIS} & \multicolumn{1}{||c}{ODS} & \multicolumn{1}{|c|}{OIS}\\
\hline
\multicolumn{1}{|c}{gPb-OWT-UCM~\cite{arbelaez2011contour}} & \multicolumn{1}{|c}{$0.59$} & \multicolumn{1}{|c}{$0.65$} & \multicolumn{1}{||c}{$0.81$} & \multicolumn{1}{|c}{$0.85$} & \multicolumn{1}{||c}{$1.65$} & \multicolumn{1}{|c|}{$1.47$}\\
\multicolumn{1}{|c}{ISCRA~\cite{ren2013image}} & \multicolumn{1}{|c}{$0.60$} & \multicolumn{1}{|c}{$\mathbf{0.67}$} & \multicolumn{1}{||c}{$0.81$} & \multicolumn{1}{|c}{$\mathbf{0.86}$} & \multicolumn{1}{||c}{$1.61$} & \multicolumn{1}{|c|}{$1.40$}\\
\multicolumn{1}{|c}{HOCC~\cite{kim2014image}} & \multicolumn{1}{|c}{$0.60$} & \multicolumn{1}{|c}{-} & \multicolumn{1}{||c}{$0.81$} & \multicolumn{1}{|c}{-} & \multicolumn{1}{||c}{$1.74$} & \multicolumn{1}{|c|}{-}\\
\multicolumn{1}{|c}{MCG~\cite{arbelaez2014multiscale}} & \multicolumn{1}{|c}{$\mathbf{0.61}$} & \multicolumn{1}{|c}{$\mathbf{0.67}$} & \multicolumn{1}{||c}{$0.81$} & \multicolumn{1}{|c}{$\mathbf{0.86}$} & \multicolumn{1}{||c}{$\mathbf{1.55}$} & \multicolumn{1}{|c|}{$\mathbf{1.37}$}\\
\multicolumn{1}{|c}{HMT} & \multicolumn{1}{|c}{$\mathbf{0.61}$} & \multicolumn{1}{|c}{$\mathbf{0.67}$} & \multicolumn{1}{||c}{$\mathbf{0.82}$} & \multicolumn{1}{|c}{$\mathbf{0.86}$} & \multicolumn{1}{||c}{$1.58$} & \multicolumn{1}{|c|}{$1.40$}\\
\hline
\end{tabular}
}\\
\subfloat[\label{subtab:exp_others_bsds500}]{
\begin{tabular}{ccccccc}
\cline{2-7}
& \multicolumn{6}{|c|}{\textbf{BSDS500}}\\
\cline{2-7}
& \multicolumn{2}{|c}{Covering} & \multicolumn{2}{||c}{PRI} & \multicolumn{2}{||c|}{VI}\\
\hline
\multicolumn{1}{|c}{Method} & \multicolumn{1}{|c}{ODS} & \multicolumn{1}{|c}{OIS} & \multicolumn{1}{||c}{ODS} & \multicolumn{1}{|c}{OIS} & \multicolumn{1}{||c}{ODS} & \multicolumn{1}{|c|}{OIS}\\
\hline
\multicolumn{1}{|c}{gPb-OWT-UCM~\cite{arbelaez2011contour}} & \multicolumn{1}{|c}{$0.59$} & \multicolumn{1}{|c}{$0.65$} & \multicolumn{1}{||c}{$0.83$} & \multicolumn{1}{|c}{$0.86$} & \multicolumn{1}{||c}{$1.69$} & \multicolumn{1}{|c|}{$1.48$}\\
\multicolumn{1}{|c}{ISCRA~\cite{ren2013image}} & \multicolumn{1}{|c}{$0.59$} & \multicolumn{1}{|c}{$0.66$} & \multicolumn{1}{||c}{$0.82$} & \multicolumn{1}{|c}{$0.86$} & \multicolumn{1}{||c}{$1.60$} & \multicolumn{1}{|c|}{$1.42$}\\
\multicolumn{1}{|c}{GALA~\cite{nunez2013machine}} & \multicolumn{1}{|c}{$0.61$} & \multicolumn{1}{|c}{$0.67$} & \multicolumn{1}{||c}{$\mathbf{0.84}$} & \multicolumn{1}{|c}{$0.86$} & \multicolumn{1}{||c}{$1.56$} & \multicolumn{1}{|c|}{$\mathbf{1.36}$}\\
\multicolumn{1}{|c}{HOCC~\cite{kim2014image}} & \multicolumn{1}{|c}{$0.60$} & \multicolumn{1}{|c}{-} & \multicolumn{1}{||c}{$0.83$} & \multicolumn{1}{|c}{-} & \multicolumn{1}{||c}{$1.79$} & \multicolumn{1}{|c|}{-}\\
\multicolumn{1}{|c}{DC~\cite{donoser2014discrete}} & \multicolumn{1}{|c}{$0.59$} & \multicolumn{1}{|c}{$0.64$} & \multicolumn{1}{||c}{$0.82$} & \multicolumn{1}{|c}{$0.85$} & \multicolumn{1}{||c}{$1.68$} & \multicolumn{1}{|c|}{$1.54$}\\
\multicolumn{1}{|c}{MCG~\cite{arbelaez2014multiscale}} & \multicolumn{1}{|c}{$0.61$} & \multicolumn{1}{|c}{$0.66$} & \multicolumn{1}{||c}{$0.83$} & \multicolumn{1}{|c}{$0.86$} & \multicolumn{1}{||c}{$1.57$} & \multicolumn{1}{|c|}{$1.39$}\\
\multicolumn{1}{|c}{PFE-mPb~\cite{yu2015piecewise}} & \multicolumn{1}{|c}{$0.62$} & \multicolumn{1}{|c}{$0.67$} & \multicolumn{1}{||c}{$\mathbf{0.84}$} & \multicolumn{1}{|c}{$0.86$} & \multicolumn{1}{||c}{$1.61$} & \multicolumn{1}{|c|}{$1.43$}\\
\multicolumn{1}{|c}{PFE-MCG~\cite{yu2015piecewise}} & \multicolumn{1}{|c}{$0.62$} & \multicolumn{1}{|c}{$\mathbf{0.68}$} & \multicolumn{1}{||c}{$\mathbf{0.84}$} & \multicolumn{1}{|c}{$\mathbf{0.87}$} & \multicolumn{1}{||c}{$1.56$} & \multicolumn{1}{|c|}{$\mathbf{1.36}$}\\
\multicolumn{1}{|c}{HMT} & \multicolumn{1}{|c}{$\mathbf{0.63}$} & \multicolumn{1}{|c}{$\mathbf{0.68}$} & \multicolumn{1}{||c}{$\mathbf{0.84}$} & \multicolumn{1}{|c}{$\mathbf{0.87}$} & \multicolumn{1}{||c}{$\mathbf{1.53}$} & \multicolumn{1}{|c|}{$1.38$}\\
\hline
\end{tabular}
}\\
\subfloat[\label{subtab:exp_others_msrc}]{
\begin{tabular}{ccccccc}
\cline{2-7}
& \multicolumn{6}{|c|}{\textbf{MSRC}}\\
\cline{2-7}
& \multicolumn{2}{|c}{Covering} & \multicolumn{2}{||c}{PRI} & \multicolumn{2}{||c|}{VI}\\
\hline
\multicolumn{1}{|c}{Method} & \multicolumn{1}{|c}{ODS} & \multicolumn{1}{|c}{OIS} & \multicolumn{1}{||c}{ODS} & \multicolumn{1}{|c}{OIS} & \multicolumn{1}{||c}{ODS} & \multicolumn{1}{|c|}{OIS}\\
\hline
\multicolumn{1}{|c}{gPb-OWT-UCM~\cite{arbelaez2011contour}} & \multicolumn{1}{|c}{$0.65$} & \multicolumn{1}{|c}{$0.75$} & \multicolumn{1}{||c}{$0.78$} & \multicolumn{1}{|c}{$0.85$} & \multicolumn{1}{||c}{$1.28$} & \multicolumn{1}{|c|}{$0.99$}\\
\multicolumn{1}{|c}{ISCRA~\cite{ren2013image}} & \multicolumn{1}{|c}{$\mathbf{0.67}$} & \multicolumn{1}{|c}{$0.75$} & \multicolumn{1}{||c}{$0.77$} & \multicolumn{1}{|c}{$0.85$} & \multicolumn{1}{||c}{$\mathbf{1.18}$} & \multicolumn{1}{|c|}{$1.02$}\\
\multicolumn{1}{|c}{HMT} & \multicolumn{1}{|c}{$\mathbf{0.67}$} & \multicolumn{1}{|c}{$\mathbf{0.77}$} & \multicolumn{1}{||c}{$\mathbf{0.79}$} & \multicolumn{1}{|c}{$\mathbf{0.86}$} & \multicolumn{1}{||c}{$1.23$} & \multicolumn{1}{|c|}{$\mathbf{0.93}$}\\
\hline
\end{tabular}
}\\
\subfloat[\label{subtab:exp_others_voc12}]{
\begin{tabular}{ccccccc}
\cline{2-7}
& \multicolumn{6}{|c|}{\textbf{VOC12}}\\
\cline{2-7}
& \multicolumn{2}{|c}{Covering} & \multicolumn{2}{||c}{PRI} & \multicolumn{2}{||c|}{VI}\\
\hline
\multicolumn{1}{|c}{Method} & \multicolumn{1}{|c}{ODS} & \multicolumn{1}{|c}{OIS} & \multicolumn{1}{||c}{ODS} & \multicolumn{1}{|c}{OIS} & \multicolumn{1}{||c}{ODS} & \multicolumn{1}{|c|}{OIS}\\
\hline
\multicolumn{1}{|c}{gPb-OWT-UCM~\cite{arbelaez2011contour}} & \multicolumn{1}{|c}{$0.46$} & \multicolumn{1}{|c}{$0.59$} & \multicolumn{1}{||c}{$0.76$} & \multicolumn{1}{|c}{$0.88$} & \multicolumn{1}{||c}{$0.65$} & \multicolumn{1}{|c|}{$0.50$}\\
\multicolumn{1}{|c}{ISCRA~\cite{ren2013image}} & \multicolumn{1}{|c}{$\mathbf{0.50}$} & \multicolumn{1}{|c}{$0.58$} & \multicolumn{1}{||c}{$0.69$} & \multicolumn{1}{|c}{$0.75$} & \multicolumn{1}{||c}{$1.01$} & \multicolumn{1}{|c|}{$0.93$}\\
\multicolumn{1}{|c}{HMT} & \multicolumn{1}{|c}{$0.49$} & \multicolumn{1}{|c}{$\mathbf{0.63}$} & \multicolumn{1}{||c}{$\mathbf{0.77}$} & \multicolumn{1}{|c}{$\mathbf{0.91}$} & \multicolumn{1}{||c}{$\mathbf{0.60}$} & \multicolumn{1}{|c|}{$\mathbf{0.44}$}\\
\hline
\end{tabular}
}\\
\subfloat[\label{subtab:exp_others_sbd}]{
\begin{tabular}{ccccccc}
\cline{2-7}
& \multicolumn{6}{|c|}{\textbf{SBD}}\\
\cline{2-7}
& \multicolumn{2}{|c}{Covering} & \multicolumn{2}{||c}{PRI} & \multicolumn{2}{||c|}{VI}\\
\hline
\multicolumn{1}{|c}{Method} & \multicolumn{1}{|c}{ODS} & \multicolumn{1}{|c}{OIS} & \multicolumn{1}{||c}{ODS} & \multicolumn{1}{|c}{OIS} & \multicolumn{1}{||c}{ODS} & \multicolumn{1}{|c|}{OIS}\\
\hline
\multicolumn{1}{|c}{gPb-OWT-UCM~\cite{arbelaez2011contour}} & \multicolumn{1}{|c}{$0.58$} & \multicolumn{1}{|c}{$0.64$} & \multicolumn{1}{||c}{$0.86$} & \multicolumn{1}{|c}{$0.89$} & \multicolumn{1}{||c}{$1.88$} & \multicolumn{1}{|c|}{$1.62$}\\
\multicolumn{1}{|c}{ISCRA~\cite{ren2013image}} & \multicolumn{1}{|c}{$\mathbf{0.62}$} & \multicolumn{1}{|c}{$\mathbf{0.68}$} & \multicolumn{1}{||c}{$\mathbf{0.87}$} & \multicolumn{1}{|c}{$\mathbf{0.90}$} & \multicolumn{1}{||c}{$1.73$} & \multicolumn{1}{|c|}{$1.49$}\\
\multicolumn{1}{|c}{HMT} & \multicolumn{1}{|c}{$0.61$} & \multicolumn{1}{|c}{$0.67$} & \multicolumn{1}{||c}{$0.86$} & \multicolumn{1}{|c}{$\mathbf{0.90}$} & \multicolumn{1}{||c}{$\mathbf{1.72}$} & \multicolumn{1}{|c|}{$\mathbf{1.48}$}\\
\hline
\end{tabular}
}\\
\subfloat[\label{subtab:exp_others_nyu}]{
\begin{tabular}{ccccccc}
\cline{2-7}
& \multicolumn{6}{|c|}{\textbf{NYU}}\\
\cline{2-7}
& \multicolumn{2}{|c}{Covering} & \multicolumn{2}{||c}{PRI} & \multicolumn{2}{||c|}{VI}\\
\hline
\multicolumn{1}{|c}{Method} & \multicolumn{1}{|c}{ODS} & \multicolumn{1}{|c}{OIS} & \multicolumn{1}{||c}{ODS} & \multicolumn{1}{|c}{OIS} & \multicolumn{1}{||c}{ODS} & \multicolumn{1}{|c|}{OIS}\\
\hline
\multicolumn{1}{|c}{gPb-OWT-UCM~\cite{arbelaez2011contour}} & \multicolumn{1}{|c}{$0.55$} & \multicolumn{1}{|c}{$0.60$} & \multicolumn{1}{||c}{$\mathbf{0.90}$} & \multicolumn{1}{|c}{$\mathbf{0.92}$} & \multicolumn{1}{||c}{$1.89$} & \multicolumn{1}{|c|}{$1.89$}\\
\multicolumn{1}{|c}{ISCRA~\cite{ren2013image}} & \multicolumn{1}{|c}{$\mathbf{0.57}$} & \multicolumn{1}{|c}{$\mathbf{0.62}$} & \multicolumn{1}{||c}{$\mathbf{0.90}$} & \multicolumn{1}{|c}{$\mathbf{0.92}$} & \multicolumn{1}{||c}{$\mathbf{1.82}$} & \multicolumn{1}{|c|}{$\mathbf{1.63}$}\\
\multicolumn{1}{|c}{HMT} & \multicolumn{1}{|c}{$\mathbf{0.57}$} & \multicolumn{1}{|c}{$0.61$} & \multicolumn{1}{||c}{$\mathbf{0.90}$} & \multicolumn{1}{|c}{$\mathbf{0.92}$} & \multicolumn{1}{||c}{$1.83$} & \multicolumn{1}{|c|}{$1.66$}\\
\hline
\end{tabular}
}
\end{table}

\def\figsize{0.14}
\def\fighspace{-3mm}
\afterpage{
  \begin{figure*}[!p]
    \centering
    \begin{tabular}{ccc|ccc}
      \includegraphics[width=\figsize\textwidth]{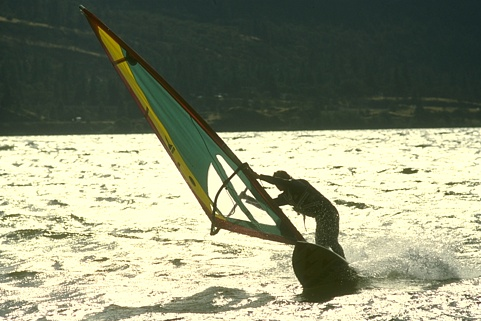}\hspace{\fighspace} &
      \includegraphics[width=\figsize\textwidth]{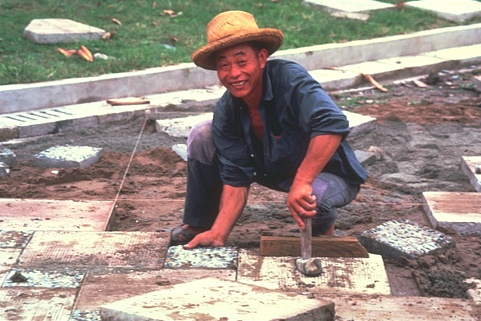}\hspace{\fighspace} &
      \includegraphics[width=\figsize\textwidth]{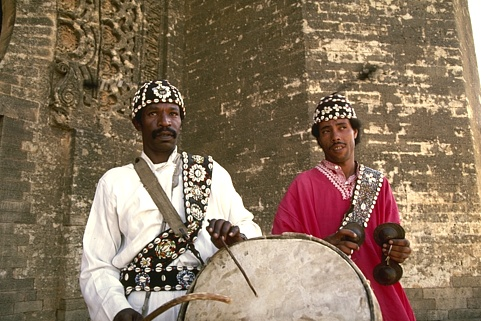} &
      \includegraphics[width=\figsize\textwidth]{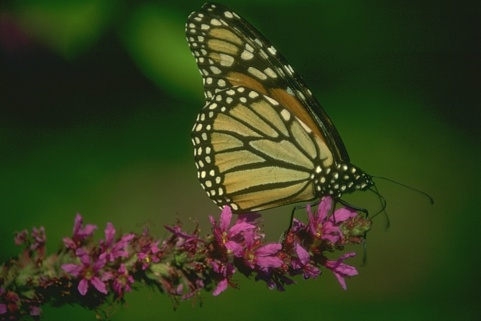}\hspace{\fighspace} &
      \includegraphics[width=\figsize\textwidth]{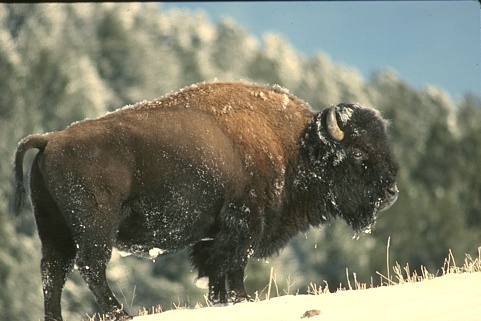}\hspace{\fighspace} &
      \includegraphics[width=\figsize\textwidth]{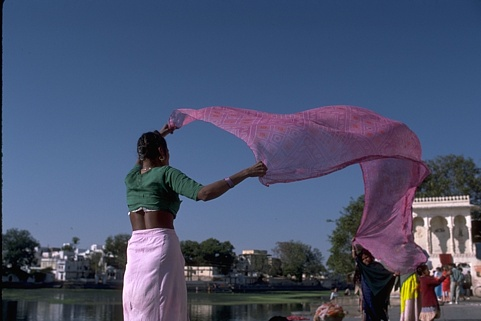}\\
      \includegraphics[width=\figsize\textwidth]{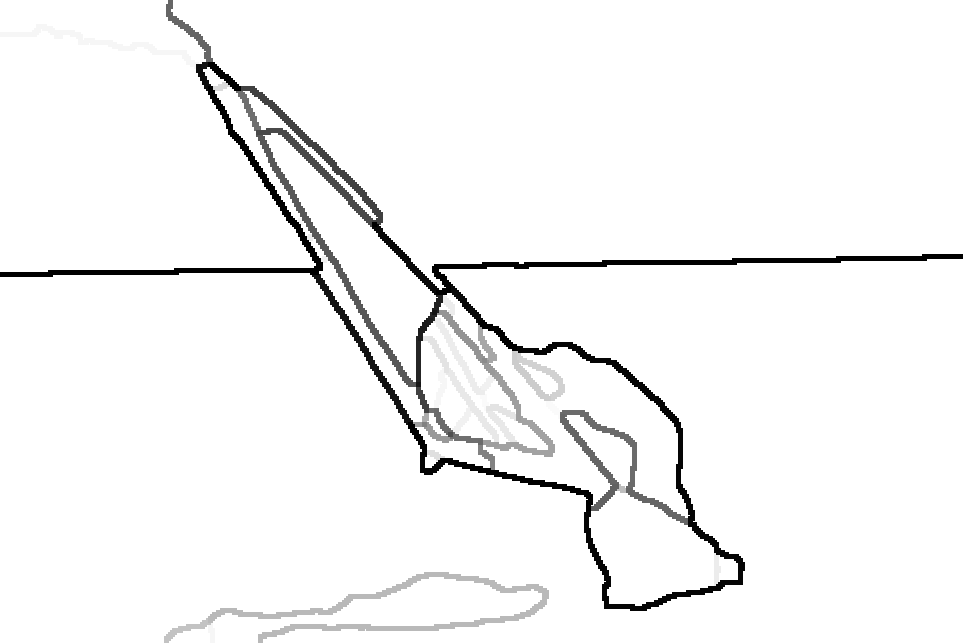}\hspace{\fighspace} &
      \includegraphics[width=\figsize\textwidth]{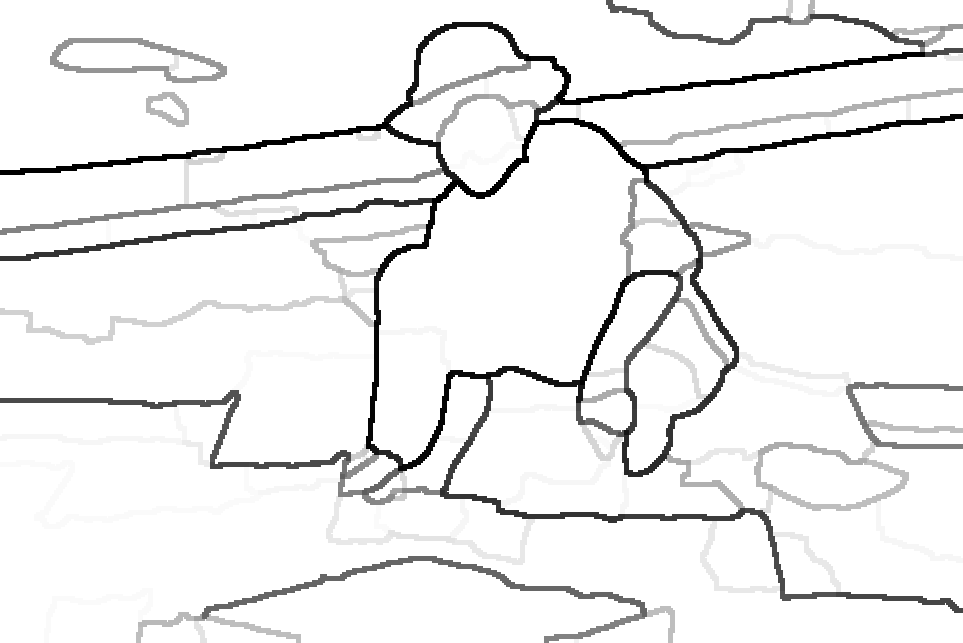}\hspace{\fighspace} &
      \includegraphics[width=\figsize\textwidth]{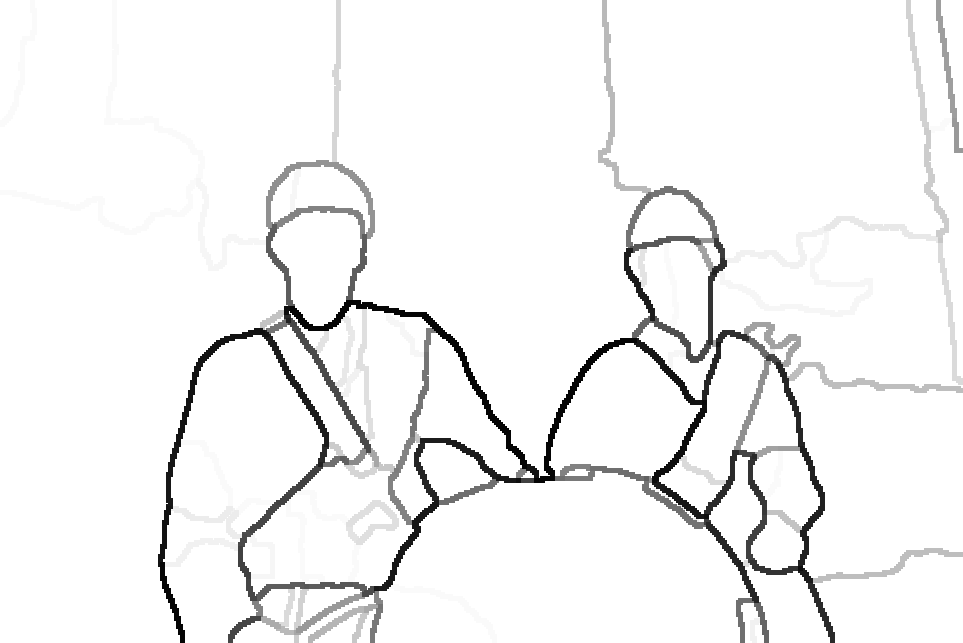} &
      \includegraphics[width=\figsize\textwidth]{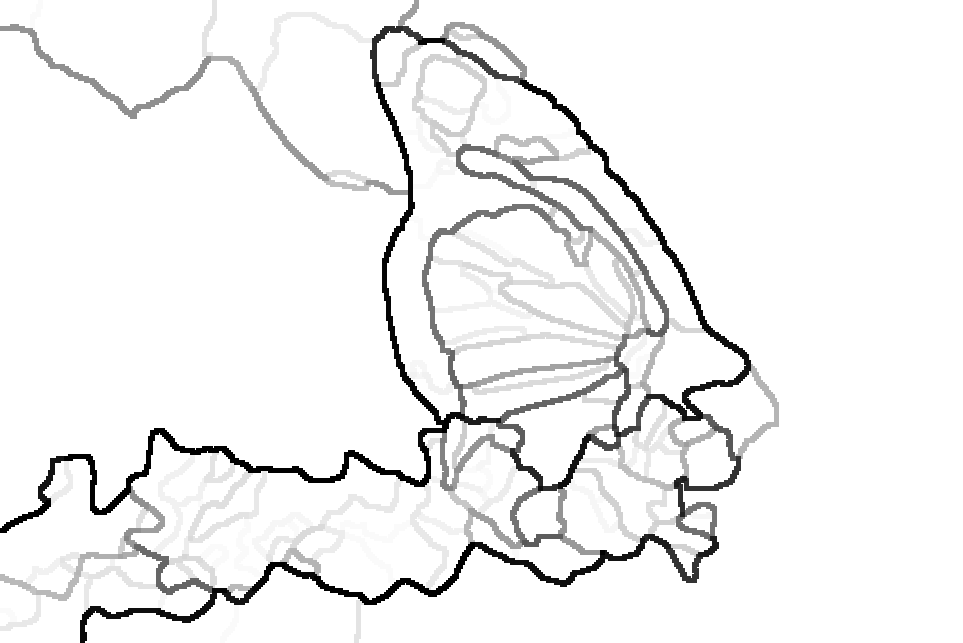}\hspace{\fighspace} &
      \includegraphics[width=\figsize\textwidth]{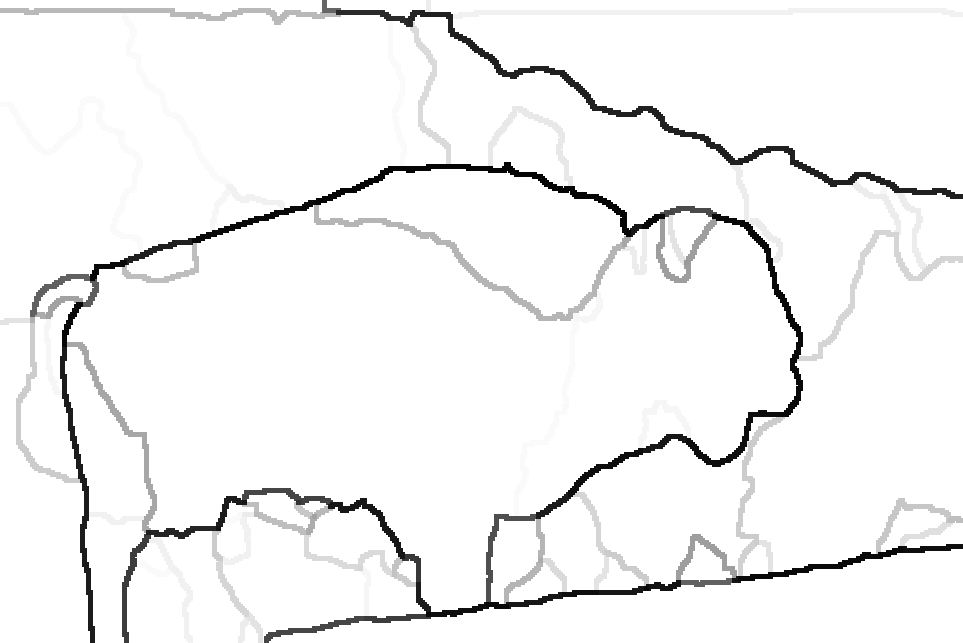}\hspace{\fighspace} &
      \includegraphics[width=\figsize\textwidth]{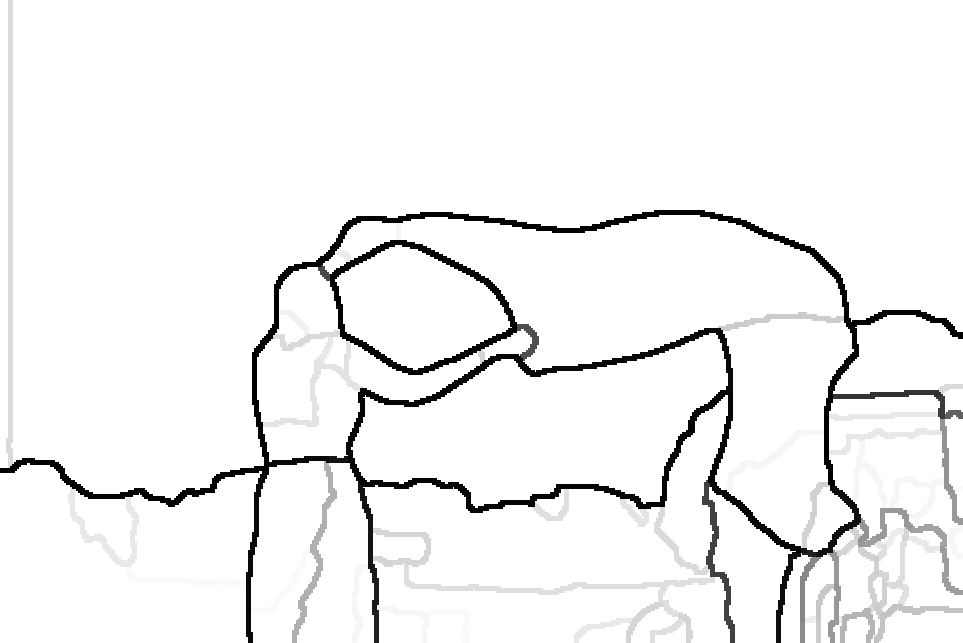}\\
      \includegraphics[width=\figsize\textwidth]{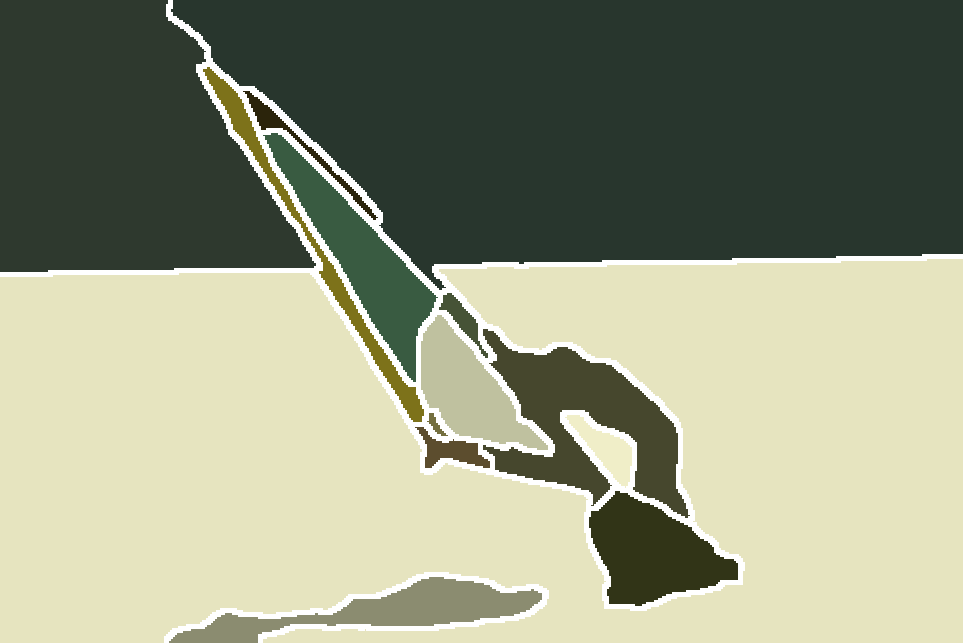}\hspace{\fighspace} &
      \includegraphics[width=\figsize\textwidth]{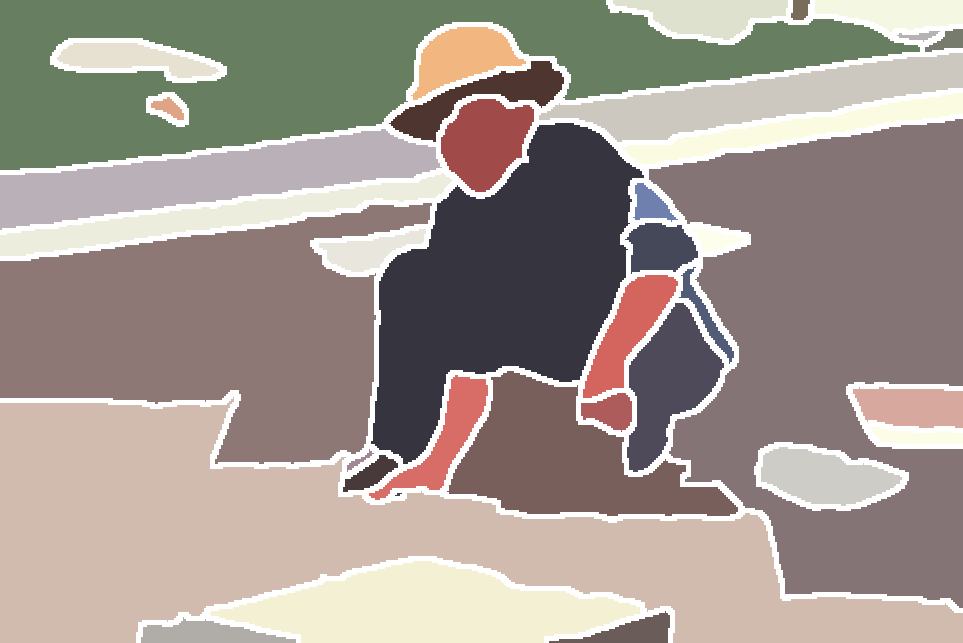}\hspace{\fighspace} &
      \includegraphics[width=\figsize\textwidth]{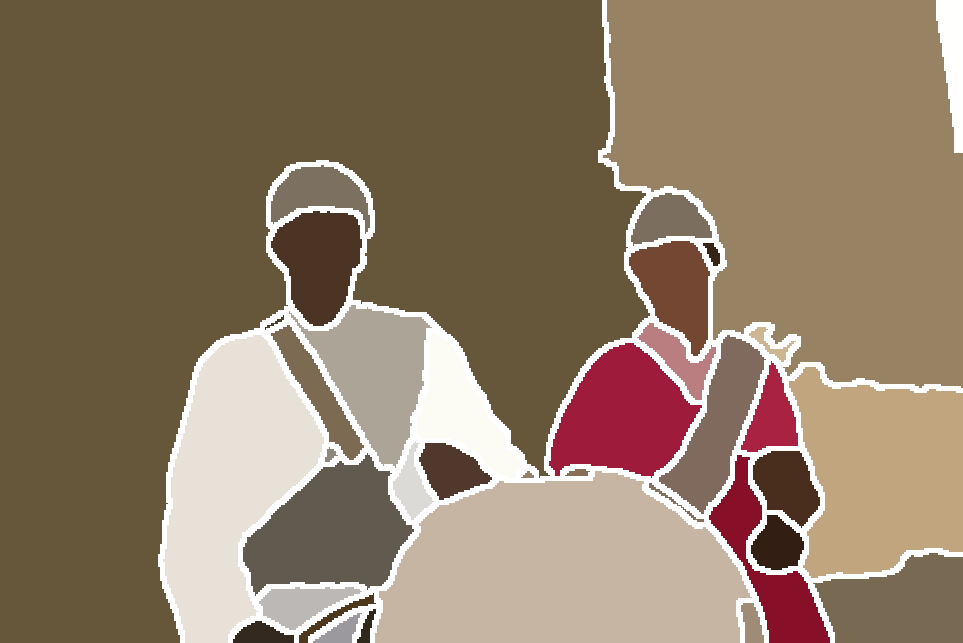} &
      \includegraphics[width=\figsize\textwidth]{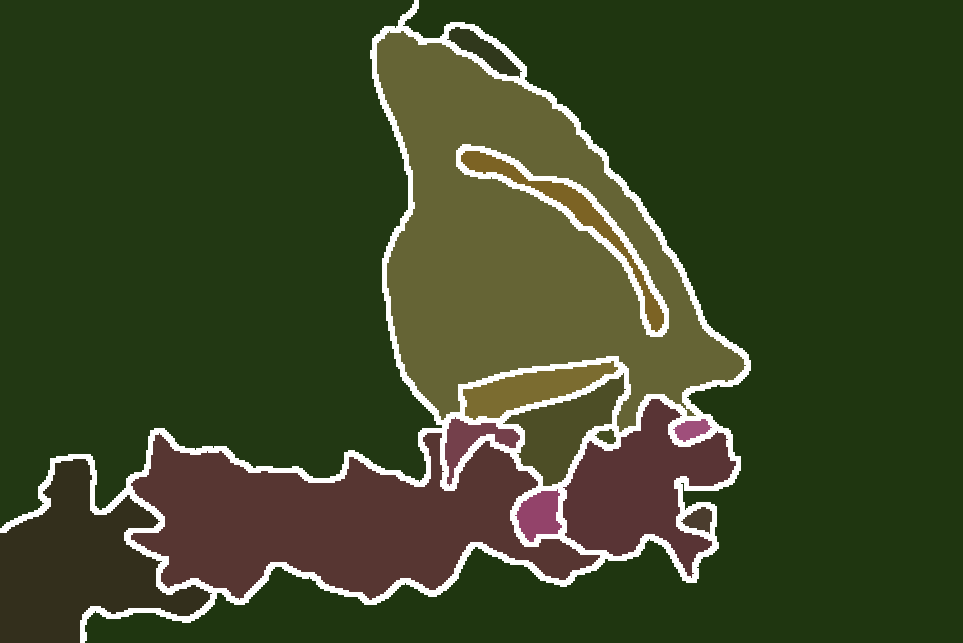}\hspace{\fighspace} &
      \includegraphics[width=\figsize\textwidth]{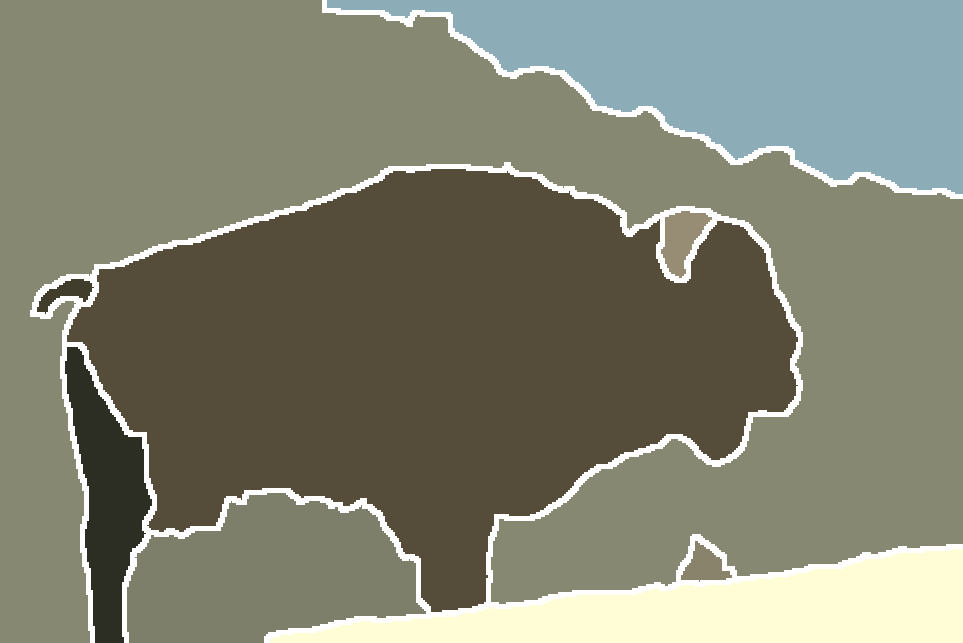}\hspace{\fighspace} &
      \includegraphics[width=\figsize\textwidth]{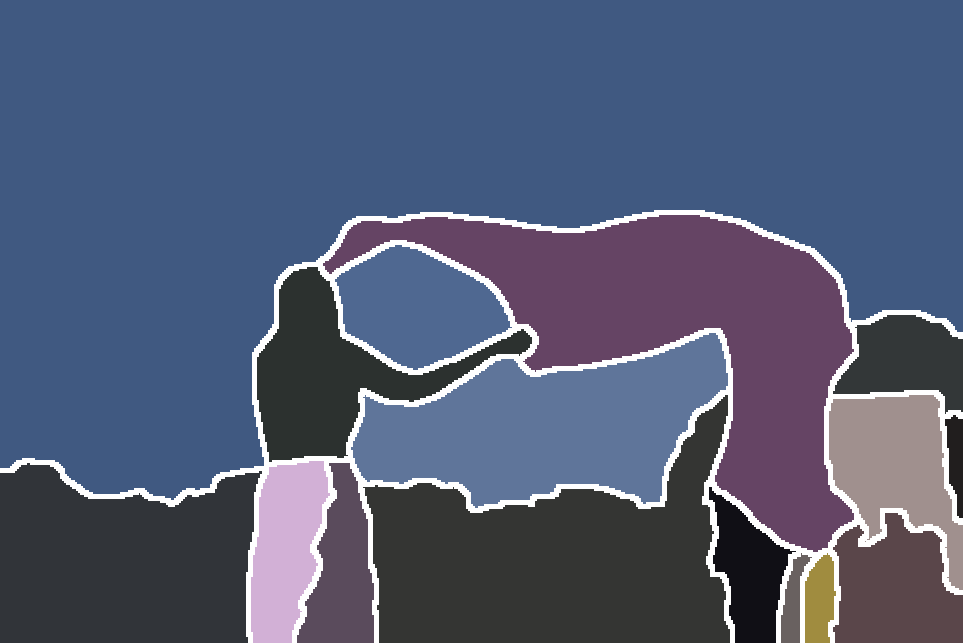}\\
      \includegraphics[width=\figsize\textwidth]{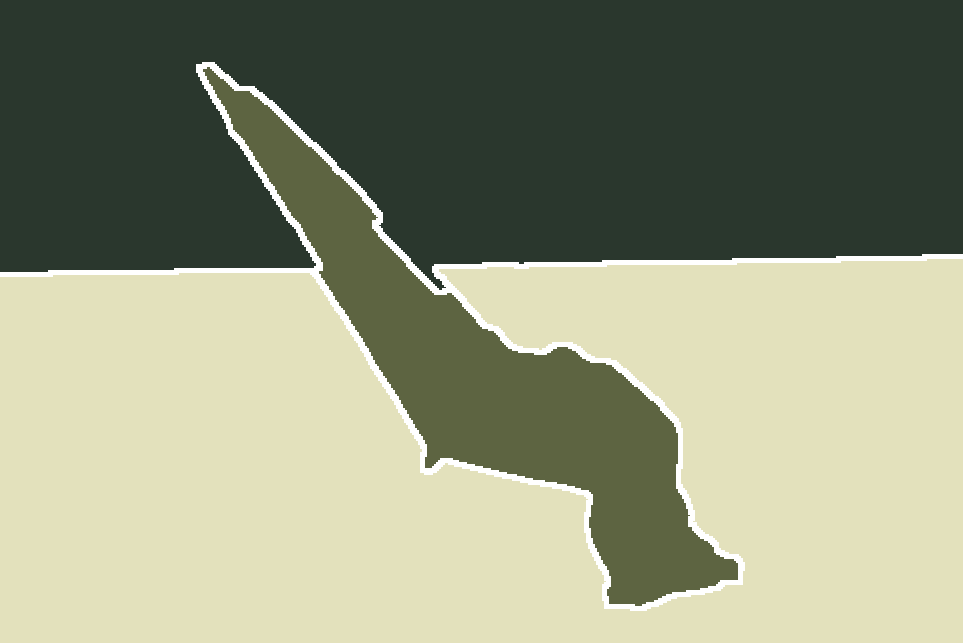}\hspace{\fighspace} &
      \includegraphics[width=\figsize\textwidth]{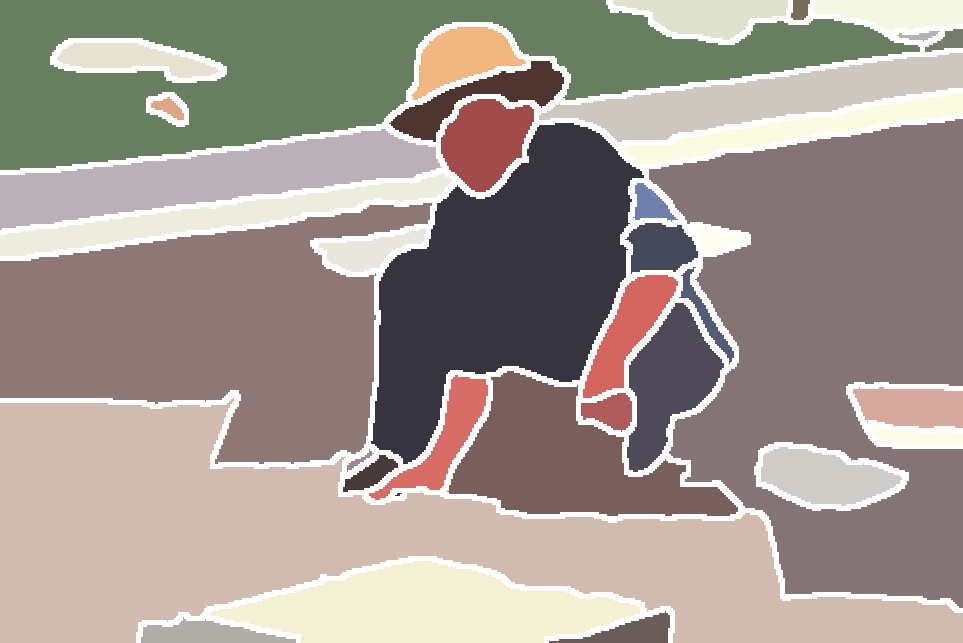}\hspace{\fighspace} &
      \includegraphics[width=\figsize\textwidth]{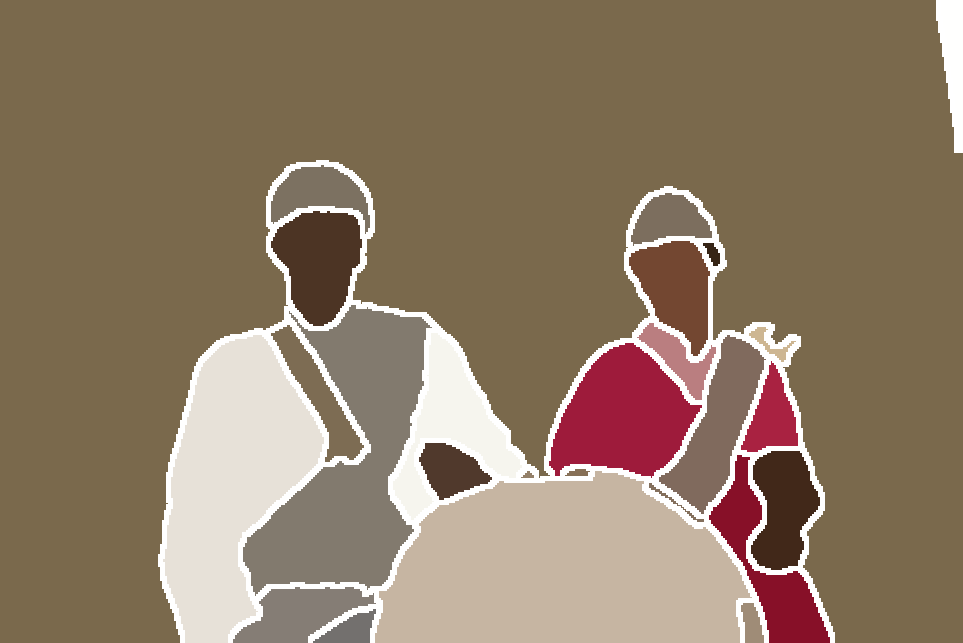} &
      \includegraphics[width=\figsize\textwidth]{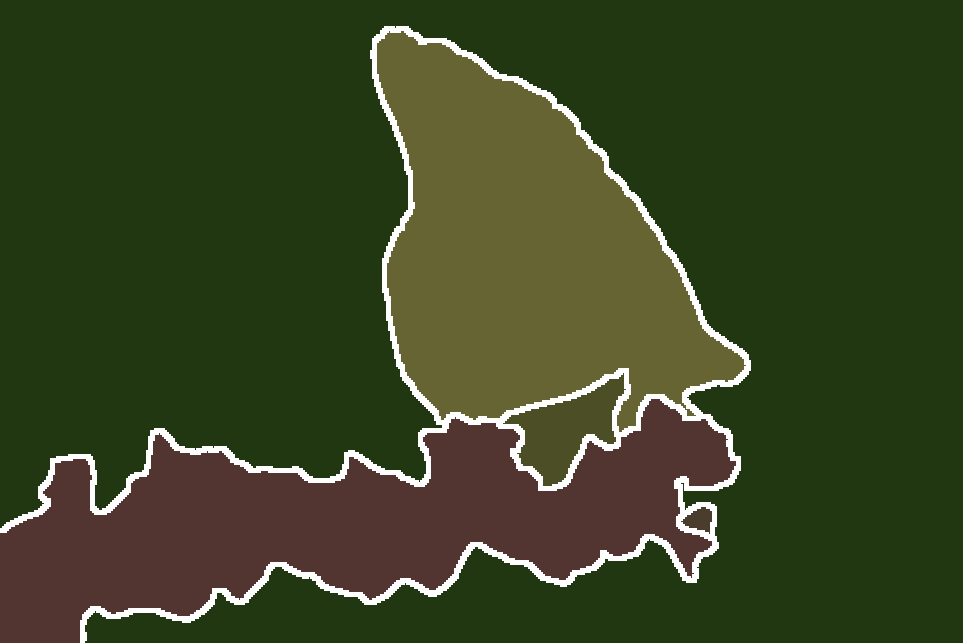}\hspace{\fighspace} &
      \includegraphics[width=\figsize\textwidth]{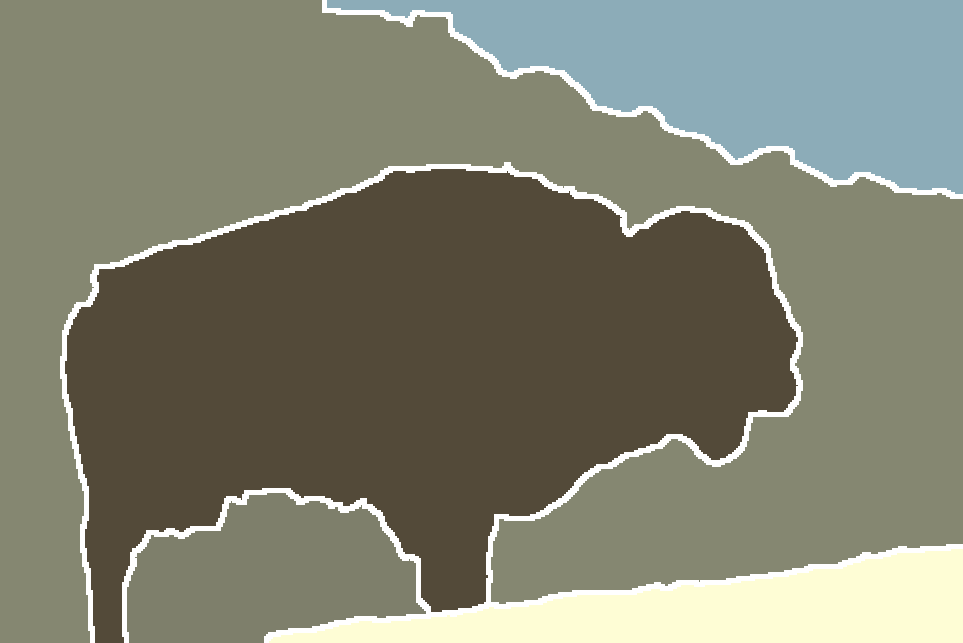}\hspace{\fighspace} &
      \includegraphics[width=\figsize\textwidth]{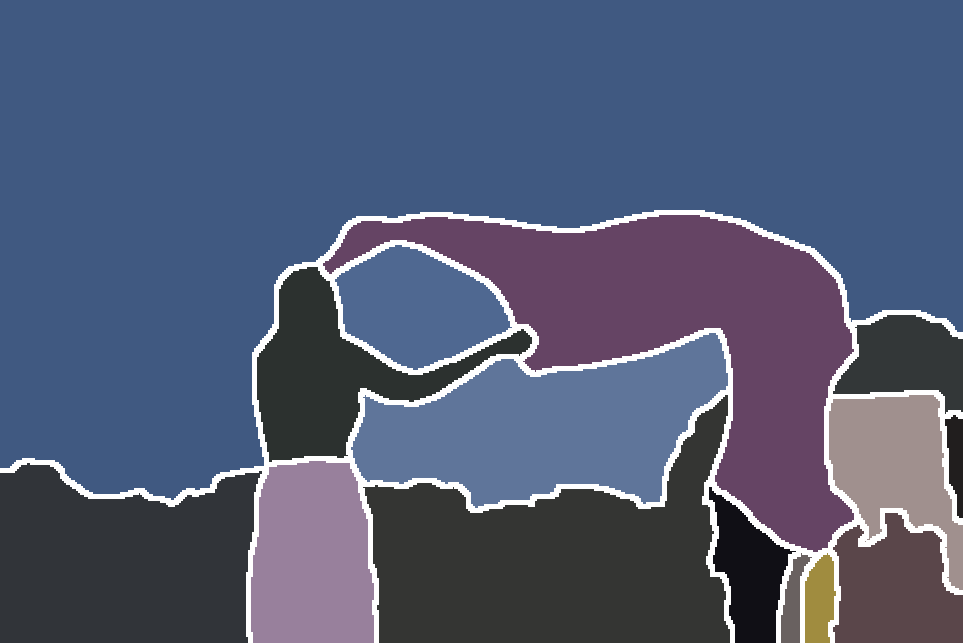}\\
      \hline
      \includegraphics[width=\figsize\textwidth]{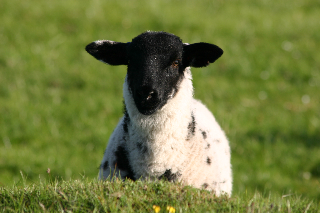}\hspace{\fighspace} &
      \includegraphics[width=\figsize\textwidth]{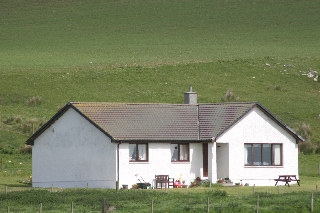}\hspace{\fighspace} &
      \includegraphics[width=\figsize\textwidth]{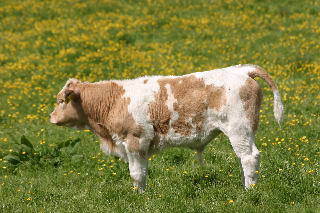} &
      \includegraphics[width=\figsize\textwidth]{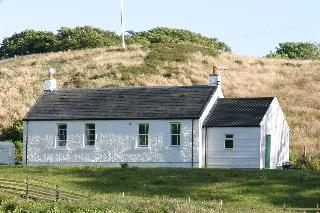}\hspace{\fighspace} &
      \includegraphics[width=\figsize\textwidth]{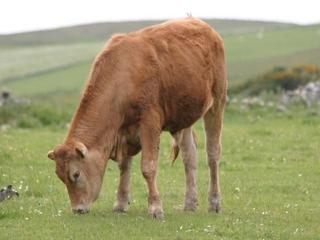}\hspace{\fighspace} &
      \includegraphics[width=\figsize\textwidth]{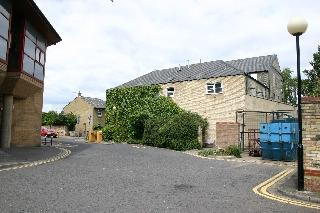}\\
      \includegraphics[width=\figsize\textwidth]{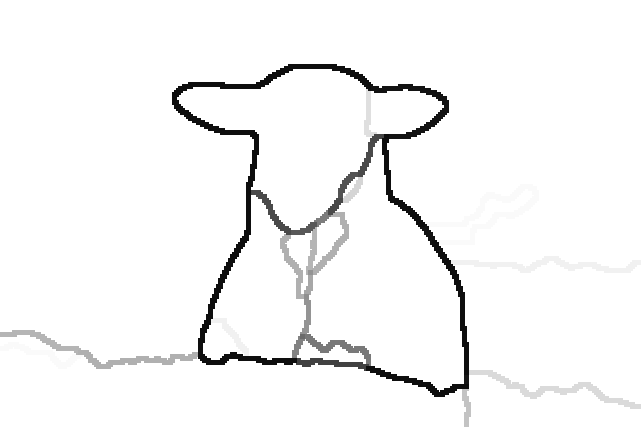}\hspace{\fighspace} &
      \includegraphics[width=\figsize\textwidth]{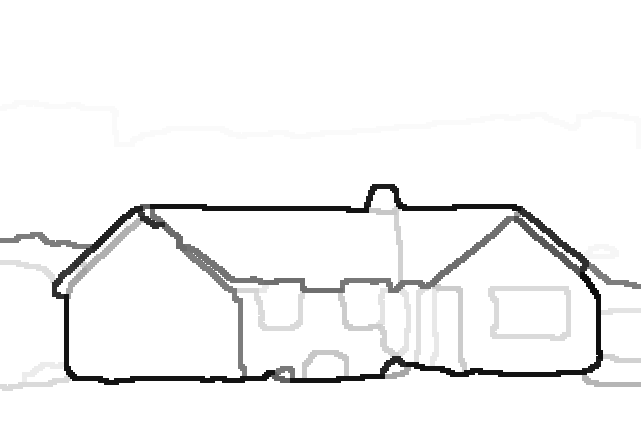}\hspace{\fighspace} &
      \includegraphics[width=\figsize\textwidth]{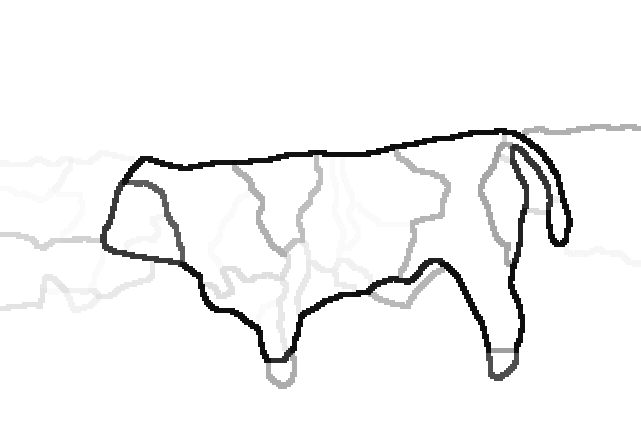} &
      \includegraphics[width=\figsize\textwidth]{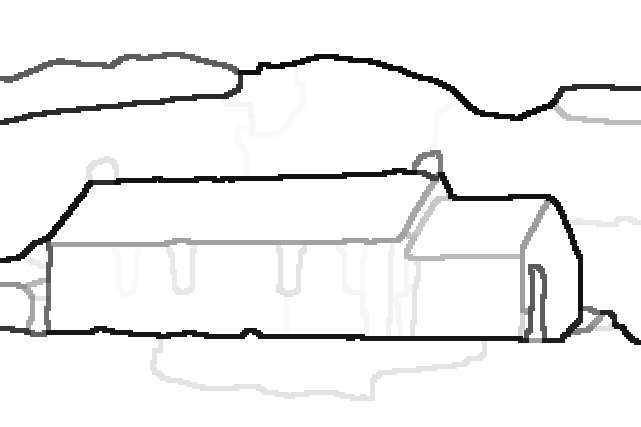}\hspace{\fighspace} &
      \includegraphics[width=\figsize\textwidth]{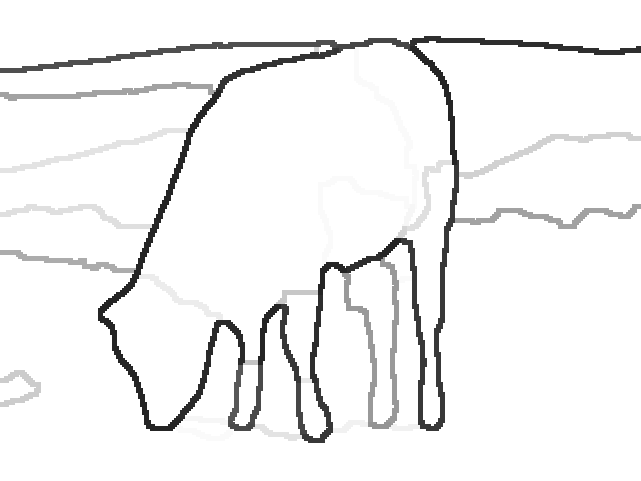}\hspace{\fighspace} &
      \includegraphics[width=\figsize\textwidth]{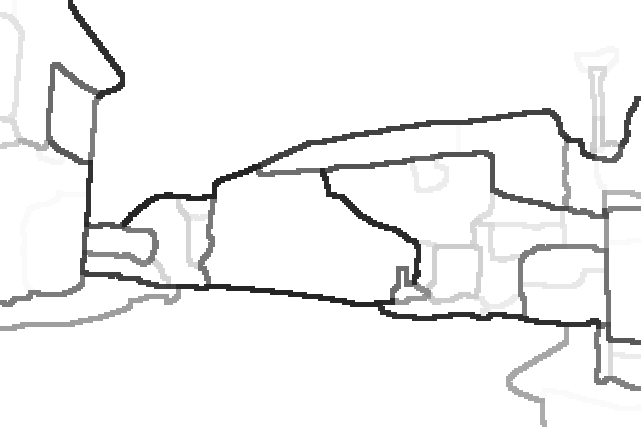}\\
      \includegraphics[width=\figsize\textwidth]{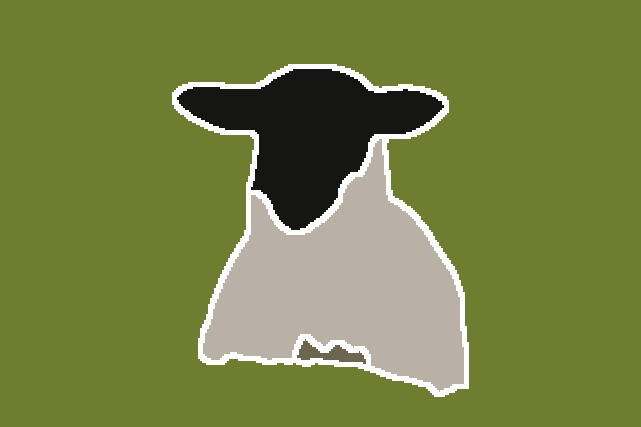}\hspace{\fighspace} &
      \includegraphics[width=\figsize\textwidth]{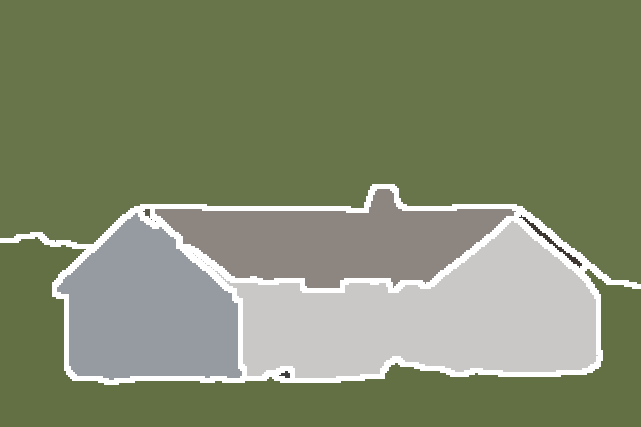}\hspace{\fighspace} &
      \includegraphics[width=\figsize\textwidth]{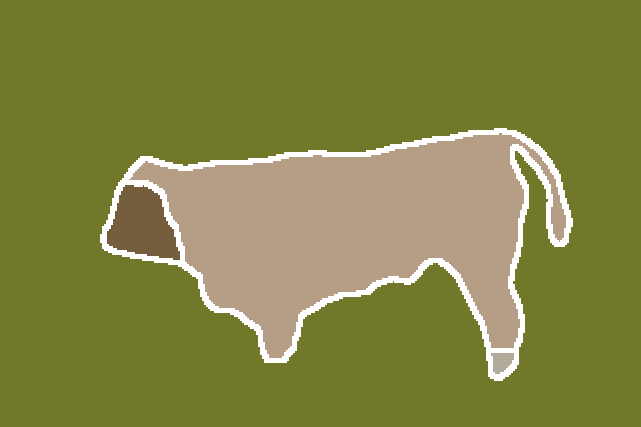} &
      \includegraphics[width=\figsize\textwidth]{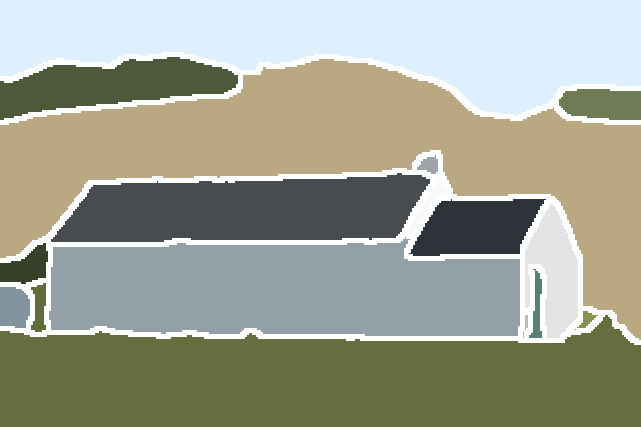}\hspace{\fighspace} &
      \includegraphics[width=\figsize\textwidth]{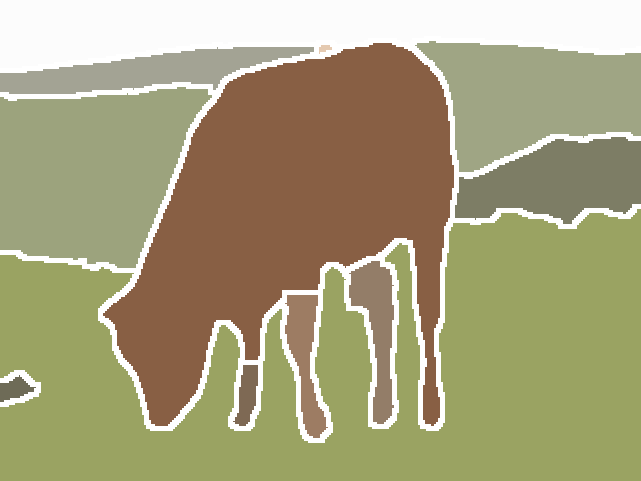}\hspace{\fighspace} &
      \includegraphics[width=\figsize\textwidth]{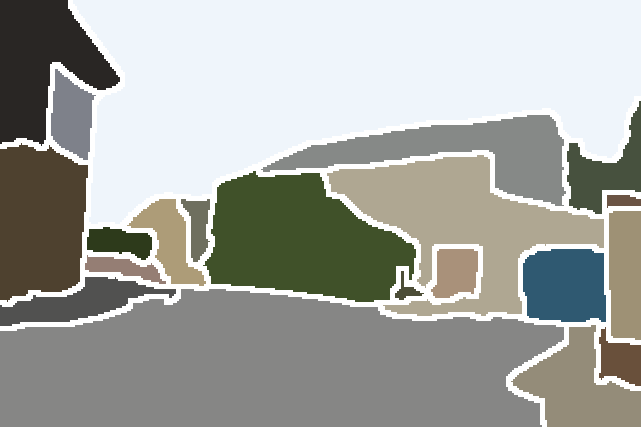}\\
      \includegraphics[width=\figsize\textwidth]{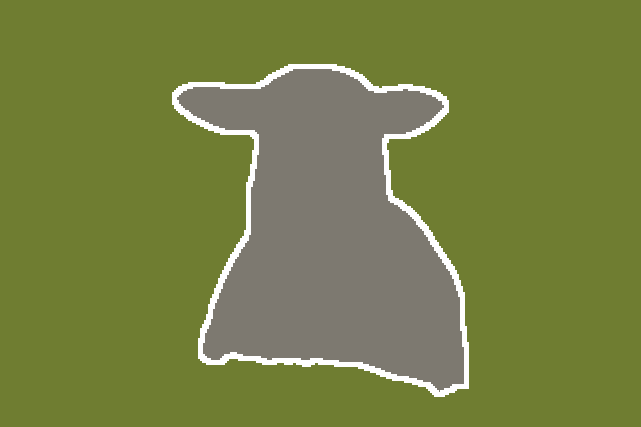}\hspace{\fighspace} &
      \includegraphics[width=\figsize\textwidth]{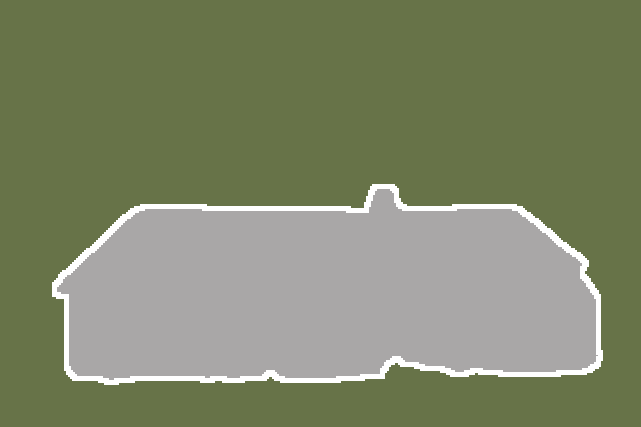}\hspace{\fighspace} &
      \includegraphics[width=\figsize\textwidth]{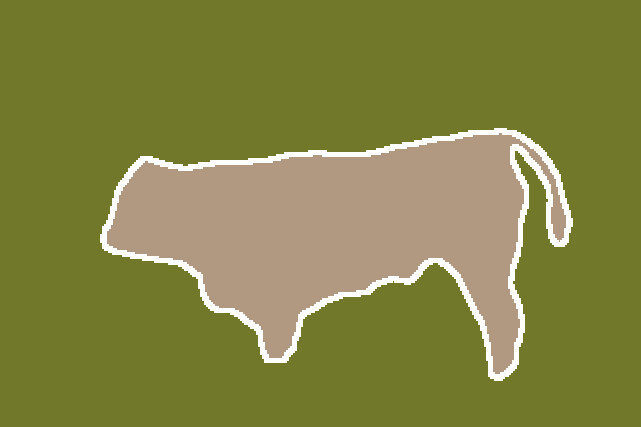} &
      \includegraphics[width=\figsize\textwidth]{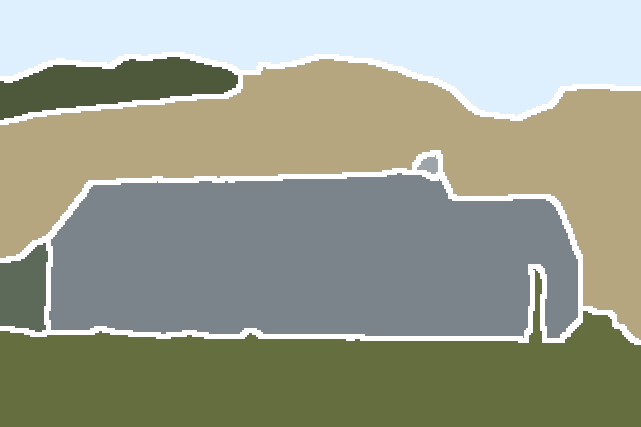}\hspace{\fighspace} &
      \includegraphics[width=\figsize\textwidth]{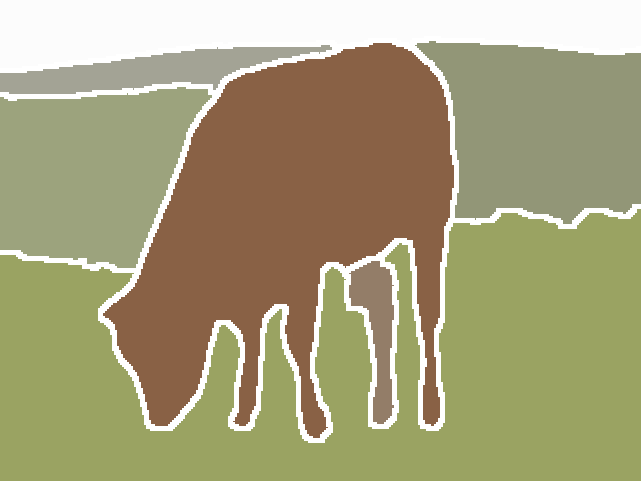}\hspace{\fighspace} &
      \includegraphics[width=\figsize\textwidth]{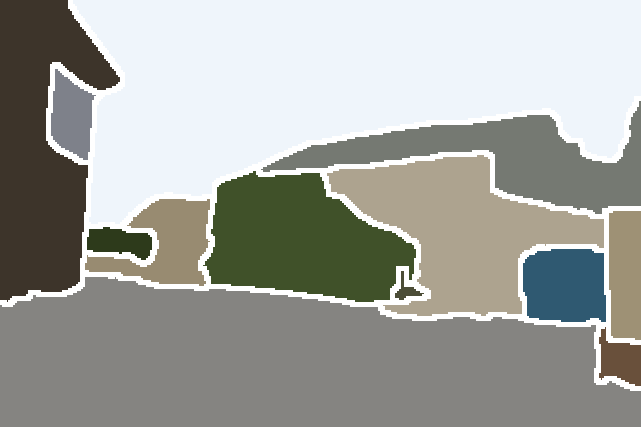}\\
      \hline
      \includegraphics[width=\figsize\textwidth]{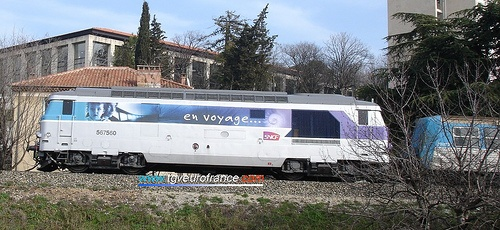}\hspace{\fighspace} &
      \includegraphics[width=\figsize\textwidth]{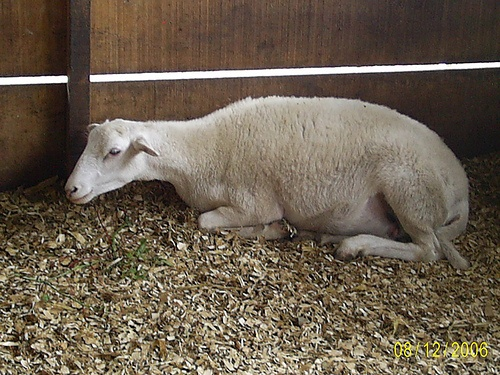}\hspace{\fighspace} &
      \includegraphics[width=\figsize\textwidth]{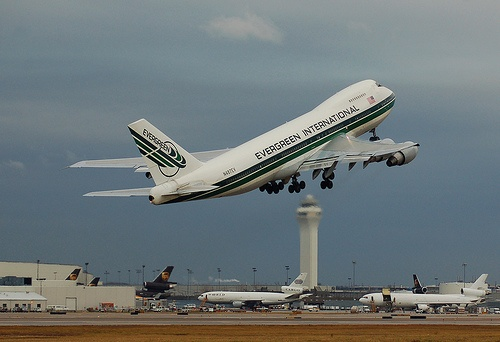} &
      \includegraphics[width=\figsize\textwidth]{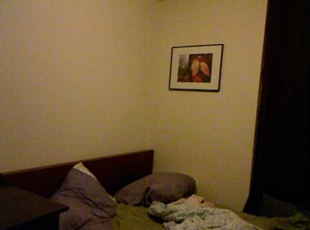}\hspace{\fighspace} &
      \includegraphics[width=\figsize\textwidth]{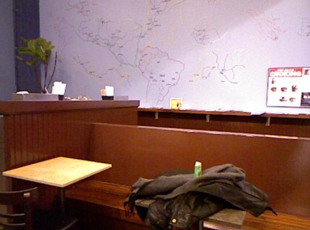}\hspace{\fighspace} &
      \includegraphics[width=\figsize\textwidth]{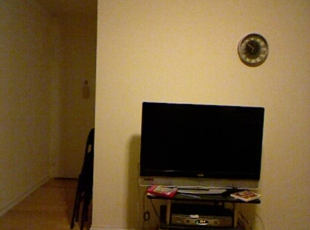}\\
      \includegraphics[width=\figsize\textwidth]{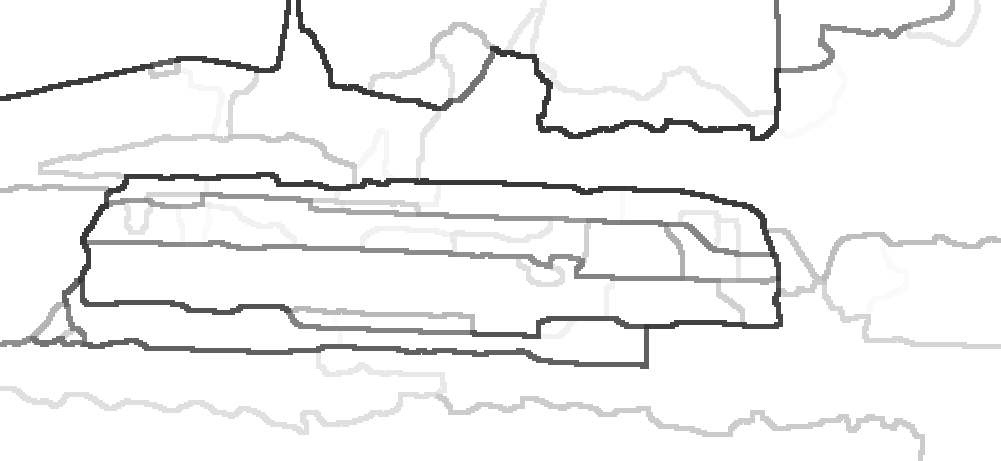}\hspace{\fighspace} &
      \includegraphics[width=\figsize\textwidth]{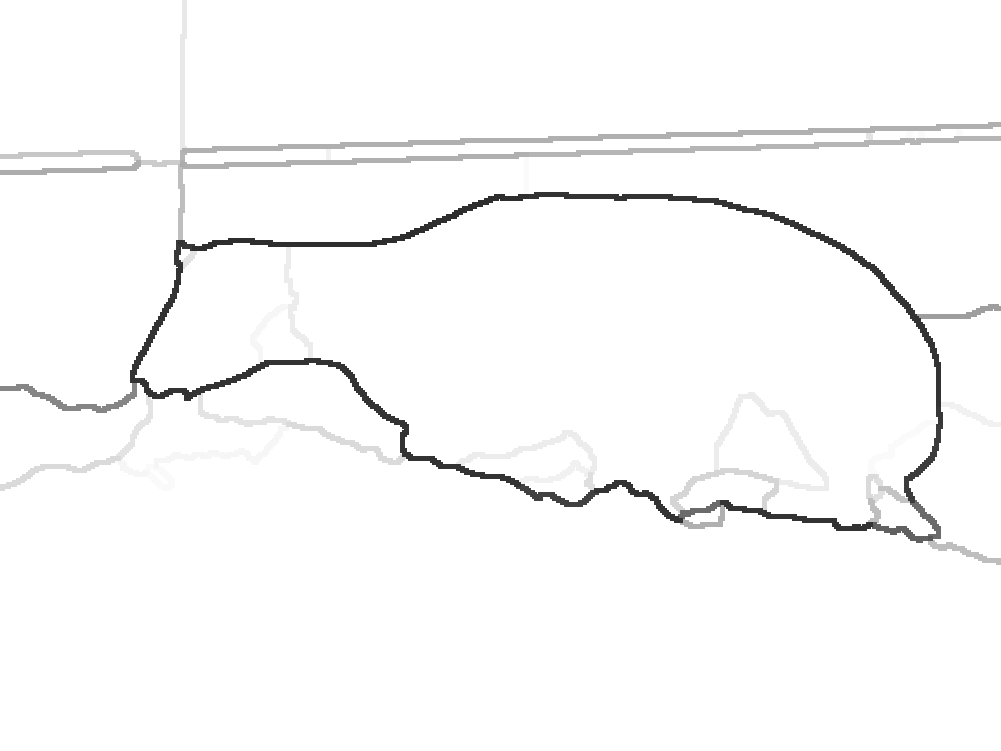}\hspace{\fighspace} &
      \includegraphics[width=\figsize\textwidth]{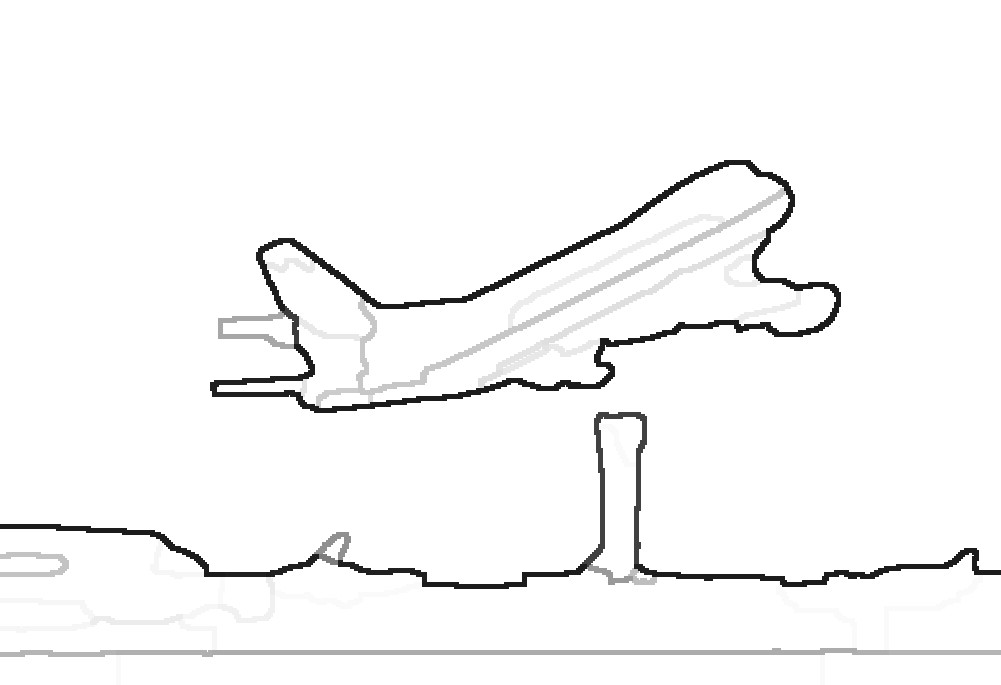} &
      \includegraphics[width=\figsize\textwidth]{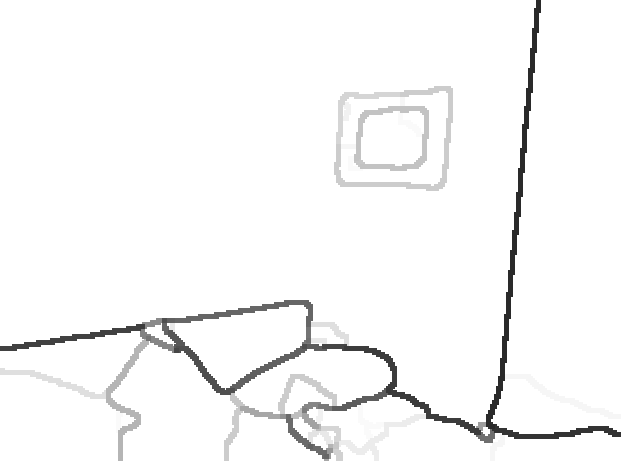}\hspace{\fighspace} &
      \includegraphics[width=\figsize\textwidth]{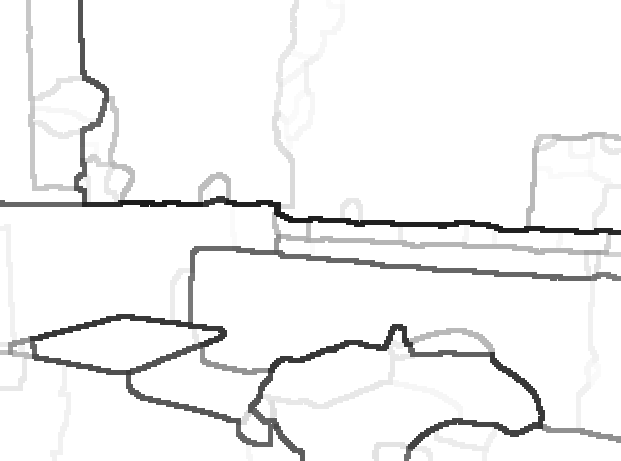}\hspace{\fighspace} &
      \includegraphics[width=\figsize\textwidth]{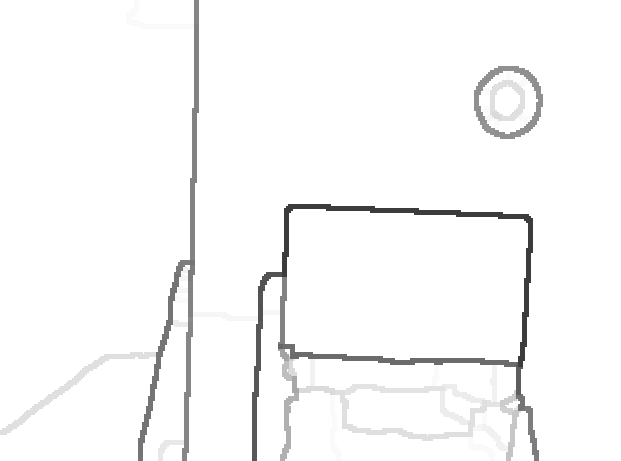}\\
      \includegraphics[width=\figsize\textwidth]{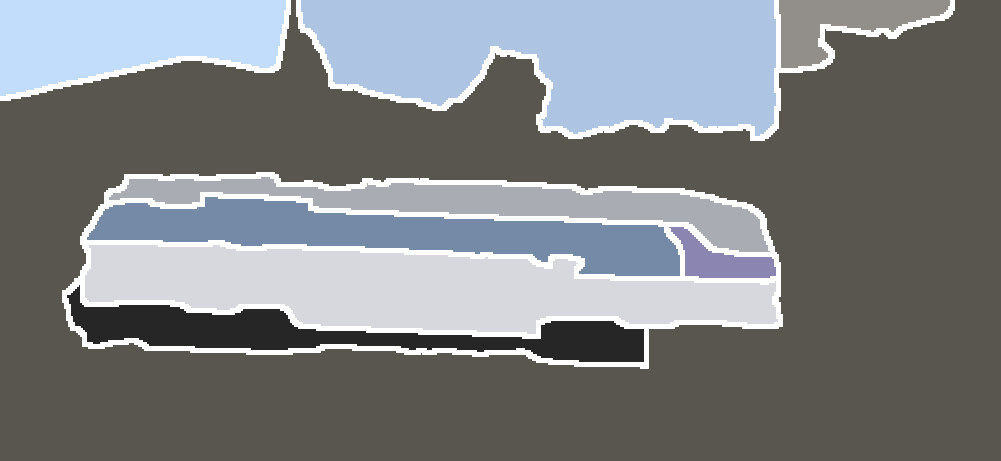}\hspace{\fighspace} &
      \includegraphics[width=\figsize\textwidth]{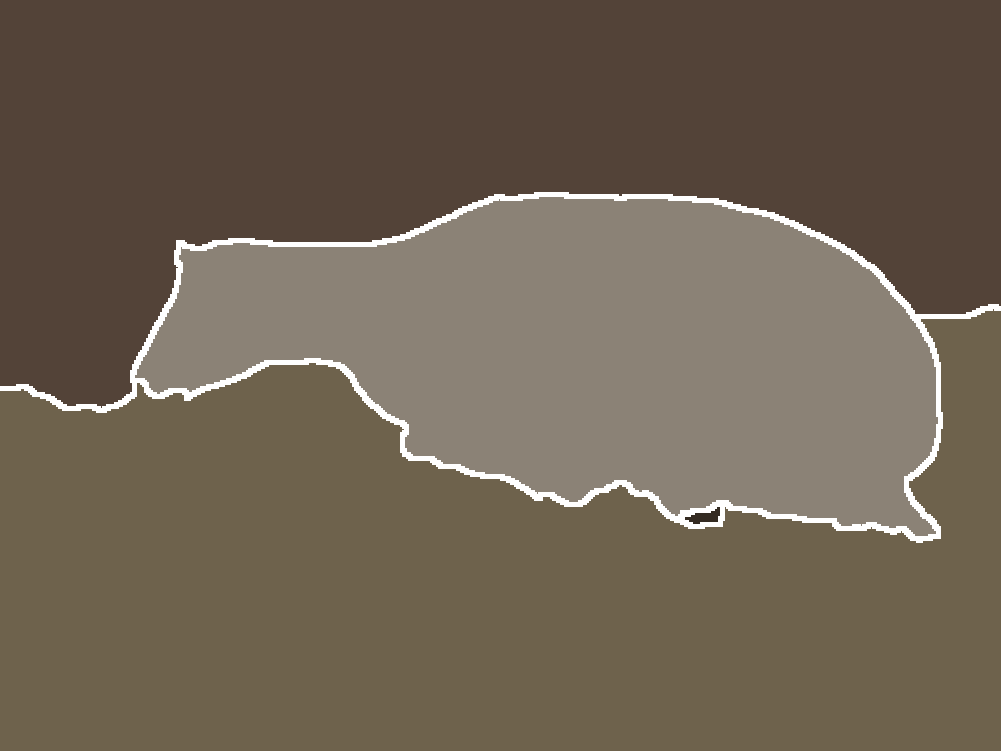}\hspace{\fighspace} &
      \includegraphics[width=\figsize\textwidth]{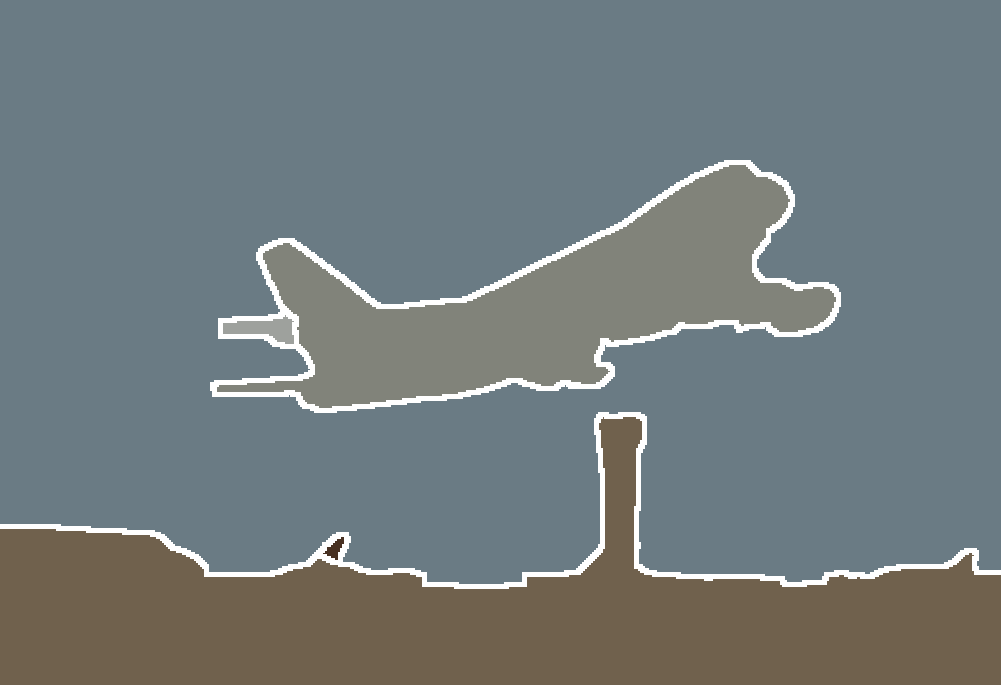} &
      \includegraphics[width=\figsize\textwidth]{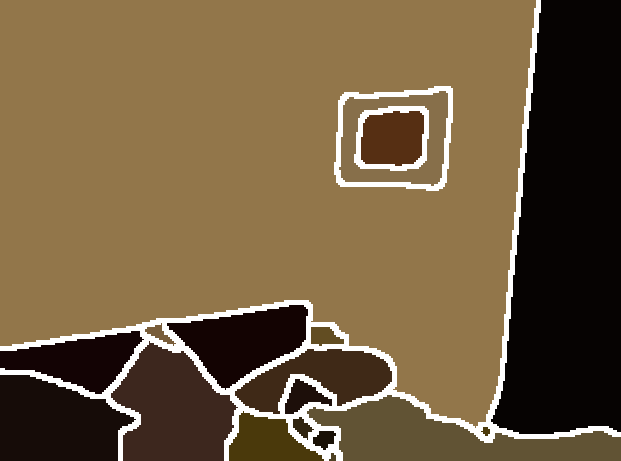}\hspace{\fighspace} &
      \includegraphics[width=\figsize\textwidth]{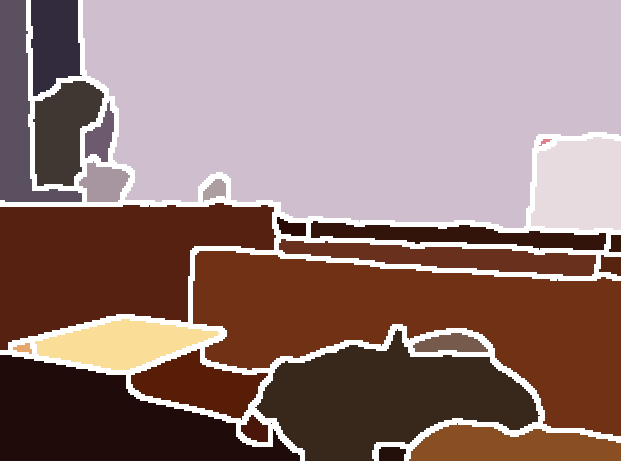}\hspace{\fighspace} &
      \includegraphics[width=\figsize\textwidth]{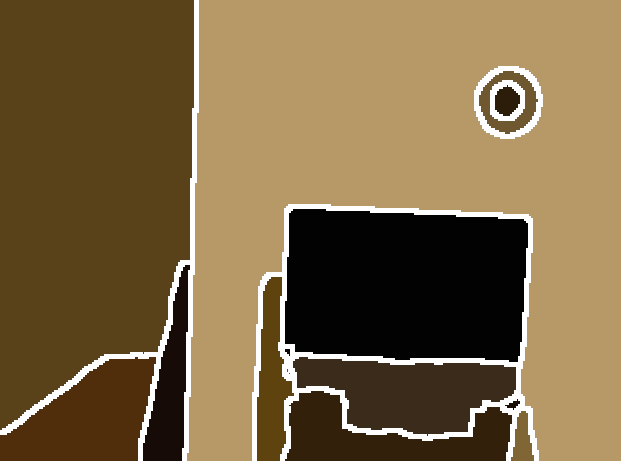}\\
      \includegraphics[width=\figsize\textwidth]{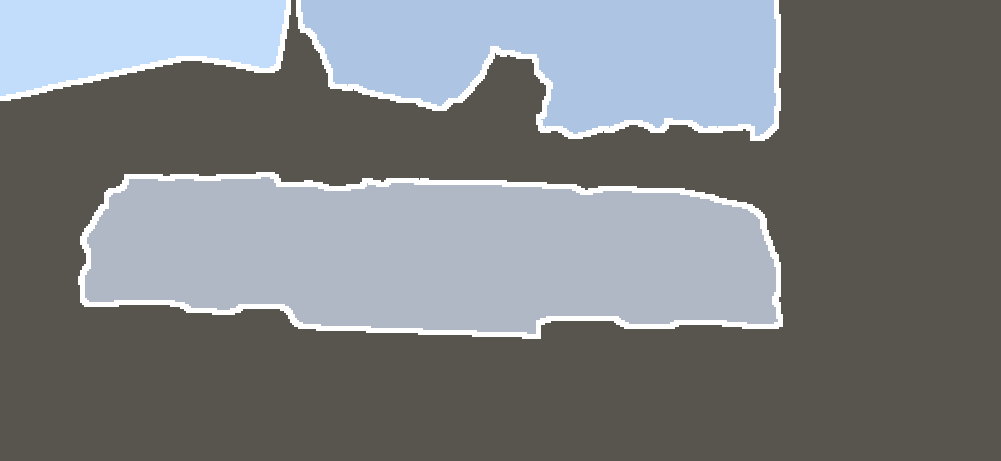}\hspace{\fighspace} &
      \includegraphics[width=\figsize\textwidth]{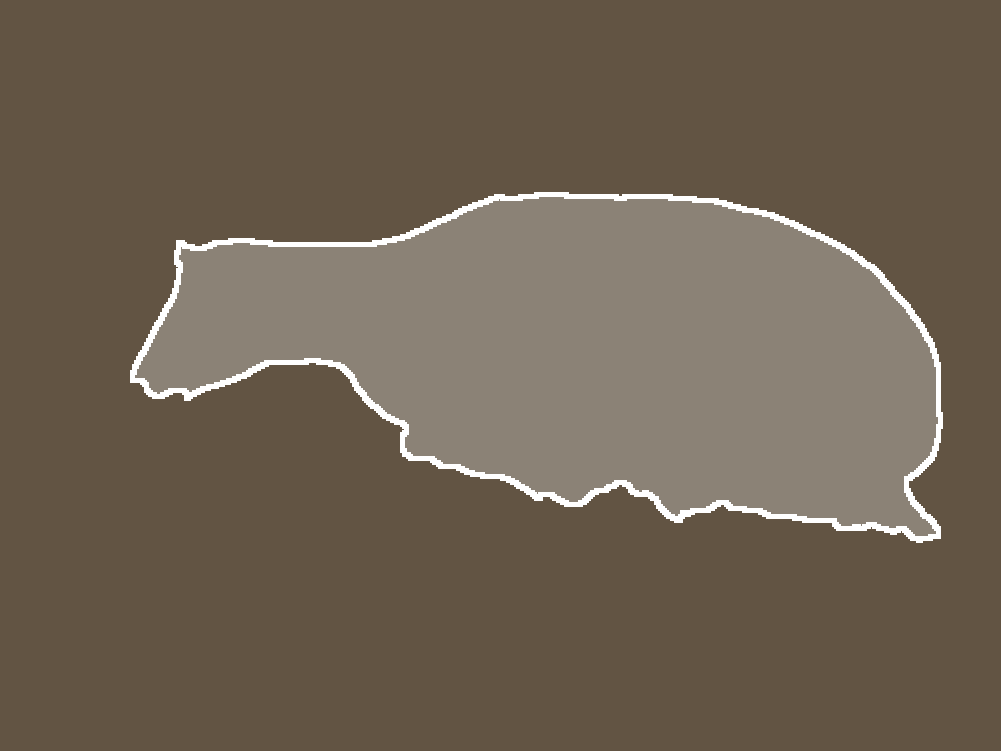}\hspace{\fighspace} &
      \includegraphics[width=\figsize\textwidth]{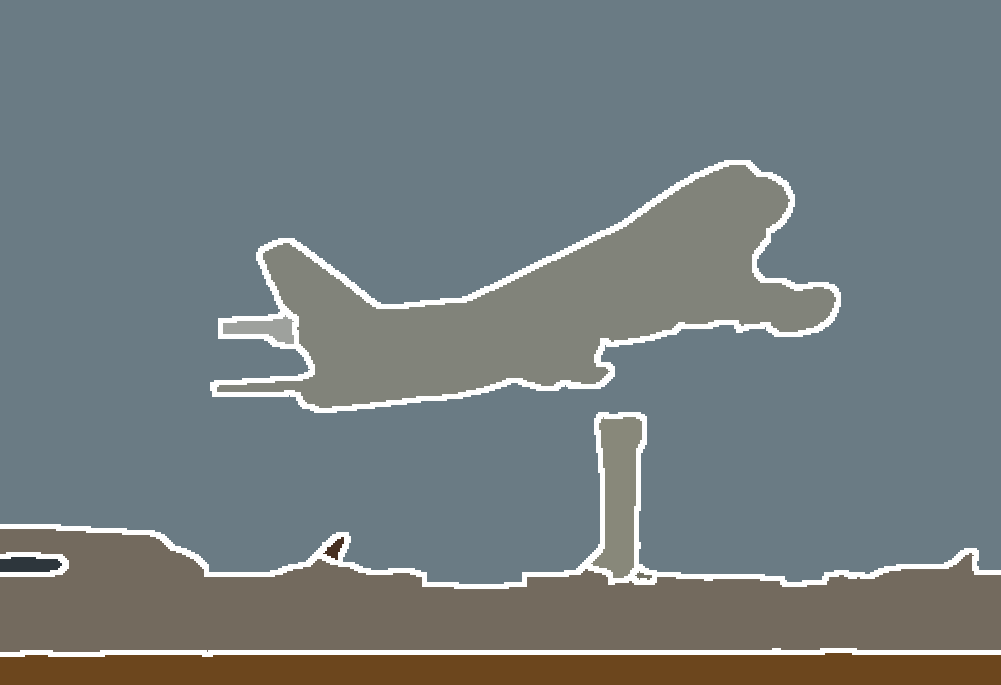} &
      \includegraphics[width=\figsize\textwidth]{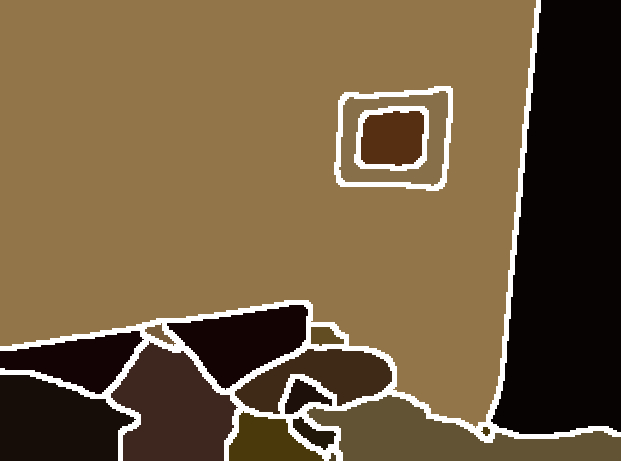}\hspace{\fighspace} &
      \includegraphics[width=\figsize\textwidth]{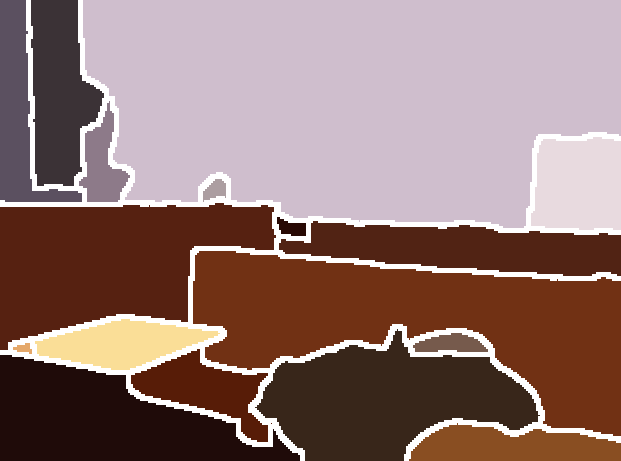}\hspace{\fighspace} &
      \includegraphics[width=\figsize\textwidth]{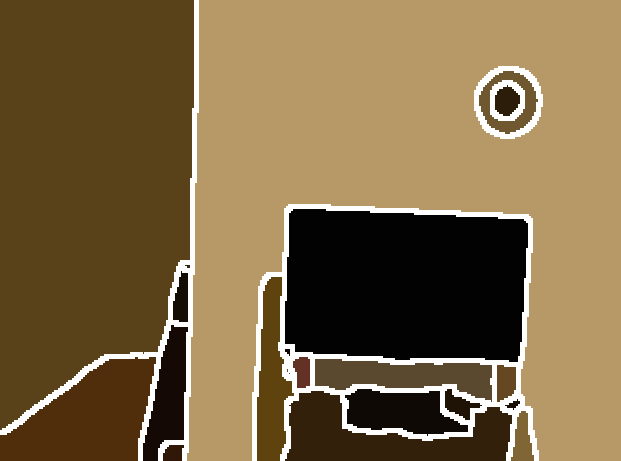}\\
    \end{tabular}
    \caption{Testing segmentation results on BSDS300 (top-left), BSDS500 (top-right), MSRC (middle-left), SBD (middle-right), VOC12 (bottom-left), and NYU (bottom-right) data set. For each image, from top to bottom: the original image, the hierarchical contour map, the ODS covering segmentation, and the OIS covering segmentation. The training uses BSDS300 training images.}\label{fig:exp_others}
  \end{figure*}
}

From Table~\ref{tab:exp_others} we can see that our method is highly competitive and outperforms very recent state-of-the-art methods on some data sets, including BSDS500, which is the most used data set for image segmentation evaluation. It is noteworthy that the generalization of our method is almost as good as ISCRA~\cite{ren2013image} by being trained only on BSDS (general natural photos) and achieving competitive results on the NYU data set (indoor scene photos). It is also worth pointing out that our hierarchical segmentation framework can be used in combination with other features that can better guide the boundary classification. For example, using the most recent piecewise flat embedding (PFE)~\cite{yu2015piecewise}, we expect the results to be further improved in a manner similar to the results from ``MCG'' to ``PFE-MCG'' on BSDS500 in Table~\ref{tab:exp_others}.

\section{Conclusion}\label{sec:conclusion}
Exhaustive search for optimal superpixel merging in image region segmentation is intractable. We propose a hierarchical image segmentation framework, namely the hierarchical merge tree model, that limits the search space to one that is induced by tree structures and thus linear with respect to the number of initial superpixels. The framework allows the use of various merging saliency heuristics and features, and its supervised nature grants its capability of learning complex conditions for merging decisions from training data without the need for parameter tuning or the dependency on any classification model. Globally optimal solutions can be efficiently found under constraints to generate final segmentations thanks to the tree structure.

We also introduce a modification to the hierarchical merge tree model that iteratively trains a new boundary classifier with accumulated samples for merge tree construction and merging probability prediction and accumulates segmentation to generate contour maps.

For further improvement, the combination of merge trees from each iteration as one single model and its global resolution can be investigated. Furthermore, it would be interesting to study the application of our method to semantic segmentation with the introduction of object-dependent prior knowledge.

\section*{Acknowledgment}
This work was supported by NSF IIS-1149299 and NIH 1R01NS075314-01. We thank Zhile Ren at Brown University for providing the ISCRA results on the testing data sets. We also thank the editor and the reviewers whose  comments greatly helped improved this paper.


%

\appendices
\section{Summary of boundary classifier features}\label{app:feat}
We use 55 features from region pairs to train the boundary classifiers, including:
\begin{enumerate}
\item Geometry (5-dimensional): Areas of two regions normalized by image area and perimeters and boundary length of two regions normalized by length of the image diagonal.
\item Boundary (4-dimensional): Means and medians of boundary pixel intensities from gPb and UCM~\cite{arbelaez2011contour}. Boundary detector gPb generates probability maps that describe how likely each pixel belong to an image boundary. UCM is the result from post-processing gPb probability maps that depicts how boundary pixels contribute to contour hierarchies in images. The boundary pixels follow the definition in~\eqref{eq:boundary_pixels}.
\item Color (24-dimensional): Absolute mean differences, $L_1$ and $\chi^2$ distances and absolute entropy differences between histograms (10-bin) of LAB and HSV components of original images.
\item Texture (8-dimensional): $L_1$ and $\chi^2$ distances between histograms of texton~\cite{malik2001contour} (64-bin) and SIFT~\cite{lowe1999object} dictionary of 256 words. The SIFT descriptors are computed densely, and $8\times8$ patches are used on gray, A, and B channel of original images.
\item Geometric context (14-dimensional): $L_1$ and $\chi^2$ distances between histograms (32-bin) of the probability maps of each of the seven geometric context labels. The geometric context labels indicate orientations of the surfaces in the images, which are predicted by a fixed pre-trained model provided by~\cite{hoiem2005geometric}.
\end{enumerate}

\section{Summary of parameters}\label{app:param}
We use the watershed algorithm for superpixel generation, for which the water level needs to be specified. In general, lowering the water level reduces under-segmentation by producing more superpixels, which gives us sets of high-precision superpixels to start with, but also increases the computation cost. We fixed the water level at $0.01$ for all five datasets (BSDS300/500, MSRC, SBD, and VOC12), except the NYU data set. For the NYU data set of indoor scene images, we observe the decrease in gPb boundary detection strength, so we lower the water level to $0.001$. We also pre-merge regions smaller than $20$ pixels to their neighboring regions with the lowest boundary barrier, i.e.\ the median of boundary detection probabilities on the boundary pixels between the two regions.

We train $255$ fully grown decision trees for the random forest boundary classifier. To train each decision tree, $70\%$ of training samples are randomly drawn and used. The number of features examined at each node is the square root of the total number of features ($\lfloor\sqrt{55}\rfloor=7$). In the experiments, the training data are usually imbalanced. The ratios between the number of positive and negative samples are sometimes considerably greater than $1$. Therefore, we assign to each class a weight reciprocal to the number of samples in the class to balance the training.

We fix the number of iterations to $T=10$ for all data sets for our iterative hierarchical merge tree model.


\ifCLASSOPTIONcaptionsoff
  \newpage
\fi



\bibliographystyle{IEEEtran}
\bibliography{refs}
\end{document}